\pdfoutput=1

\documentclass[11pt]{article}

\usepackage{amsmath,amsfonts,bm}

\def\eqref#1{equation~\ref{#1}}

\def\1{\bm{1}}

\def\vb{{\bm{b}}}

\def\vw{{\bm{w}}}

\def\mB{{\bm{B}}}

\def\mW{{\bm{W}}}
\def\mX{{\bm{X}}}
\def\mY{{\bm{Y}}}

\DeclareMathAlphabet{\mathsfit}{\encodingdefault}{\sfdefault}{m}{sl}
\SetMathAlphabet{\mathsfit}{bold}{\encodingdefault}{\sfdefault}{bx}{n}

\newcommand{\R}{\mathbb{R}}

\usepackage{EMNLP2022}

\usepackage{times}
\usepackage{latexsym}

\usepackage[T1]{fontenc}

\usepackage[utf8]{inputenc}

\usepackage{microtype}

\usepackage{inconsolata}

\usepackage{microtype}
\usepackage{graphicx}
\usepackage{subcaption}
\usepackage{booktabs} 
\usepackage{hyperref}
\usepackage{url}
\usepackage{booktabs}       
\usepackage{amsfonts}       
\usepackage{nicefrac}       
\usepackage[ruled,vlined]{algorithm2e}
\usepackage{amsmath}
\usepackage{enumitem}
\usepackage{algorithmic}
\usepackage{multirow}
\usepackage{multicol}
\usepackage{makecell}
\usepackage{subcaption}
\usepackage{eqparbox}
\usepackage{xcolor}
\usepackage{footnote}
\usepackage[flushleft]{threeparttable}
\usepackage{dsfont}

\makeatletter
\def\@xfootnote[#1]{%
  \protected@xdef\@thefnmark{#1}%
  \@footnotemark\@footnotetext}
\makeatother

\usepackage{lipsum}
\newcommand{\ie}{\textit{i.e.}}
\newcommand{\eg}{\textit{e.g.}}
\newcommand{\diag}{\text{diag}}

\newcommand{\mr}[2]{\multirow{#1}{*}{\begin{tabular}[c]{@{}c@{}}#2\end{tabular}}}

\newcommand{\PTQA}{PTQ(W\textsubscript{FT})}
\newcommand{\PTQB}{PTQ(W)+W\textsubscript{LoRA}}
\newcommand{\PTQC}{PTQ(W+W\textsubscript{LoRA})}

\newlength{\Oldarrayrulewidth}
\newcommand{\Cline}[2]{%
  \noalign{\global\setlength{\Oldarrayrulewidth}{\arrayrulewidth}}%
  \noalign{\global\setlength{\arrayrulewidth}{#1}}\cline{#2}%
  \noalign{\global\setlength{\arrayrulewidth}{\Oldarrayrulewidth}}}

\usepackage{kotex}

\title{AlphaTuning: Quantization-Aware Parameter-Efficient Adaptation\\ of Large-Scale Pre-Trained Language Models}

\author{Se Jung Kwon\textsuperscript{1}\thanks{\,\,\,Corresponding author: \texttt{sejung.kwon@navercorp.com}}, Jeonghoon Kim\textsuperscript{1}, Jeongin Bae\textsuperscript{1,4}\thanks{\,\,\,Work done while at NAVER CLOVA}, Kang Min Yoo\textsuperscript{1,2,3}, Jin-Hwa Kim\textsuperscript{2,3}, \\
        \textbf{Baeseong Park\textsuperscript{1}, Byeongwook Kim\textsuperscript{1}, Jung-Woo Ha\textsuperscript{2}, Nako Sung\textsuperscript{1} and Dongsoo Lee\textsuperscript{1}} \\
  \textsuperscript{1}NAVER CLOVA\,\,\,\,\,\,  \textsuperscript{2}NAVER AI Lab\,\,\,\,\,\, \textsuperscript{3}SNU AIIS\,\,\,\,\,\, \textsuperscript{4}KAIST 
}

\begin{document}
\maketitle
\begin{abstract}

There are growing interests in adapting large-scale language models using parameter-efficient fine-tuning methods. However, accelerating the model itself and achieving better inference efficiency through model compression has not been thoroughly explored yet.
Model compression could provide the benefits of reducing memory footprints, enabling low-precision computations, and ultimately achieving cost-effective inference.
To combine parameter-efficient adaptation and model compression, we propose AlphaTuning consisting of post-training quantization of the pre-trained language model and fine-tuning only some parts of quantized parameters for a target task.
Specifically, AlphaTuning works by employing binary-coding quantization, which factorizes the full-precision parameters into binary parameters and a separate set of scaling factors.
During the adaptation phase, the binary values are frozen for all tasks, while the scaling factors are fine-tuned for the downstream task.
We demonstrate that AlphaTuning, when applied to GPT-2 and OPT, performs competitively with full fine-tuning on a variety of downstream tasks while achieving >10$\times$ compression ratio under 4-bit quantization and >1,000$\times$ reduction in the number of trainable parameters.
\end{abstract}

\section{Introduction}


Self-supervised learning facilitates the increased number of parameters to construct pre-trained language models (PLMs) (\eg, \citet{gpt3, BERT}).
We expect the continuation of model scaling of the PLMs, especially for the Transformers \cite{transformer_original}, because their general capability follows the power-law in parameter size, exhibiting \textit{"the high-level predictability and appearance of useful capabilities"}~\cite{ganguli2022predictability}.
Therefore, the Transformer-based PLMs have been studied with great enthusiasm for various applications including natural language processing \citep{BERT, gpt2, gpt3, megatron-turing, gopher, chinchilla, palm, hyperclova}, automatic speech recognition \citep{wav2vec2}, and computer vision ~\citep{he2022masked,xie2022simmim}.


Despite the impressive zero or few-shot learning performance of PLMs, additional \textit{adaptation} steps (\eg, fine-tuning on a target task) are required to further enhance performance on downstream tasks.
Since each downstream task needs to load/store independent adaptation outcomes, if we aim to deploy multiple instances of distinct tasks, adapting PLMs with limited trainable parameters is crucial for the efficient deployment~\cite{li2018general}.
Thus, various parameter-efficient adaptation techniques, such as adapter modules \citep{adapter}, low-rank adaptation \citep{lora}, prefix-tuning \citep{prefixtuning}, prompt tuning \citep{promptsurvey, discreteprompt}, and p-tuning \citep{p-tuning}, are proposed.

\begin{figure*}[t]
    \centering
    \includegraphics[width=0.9\linewidth]{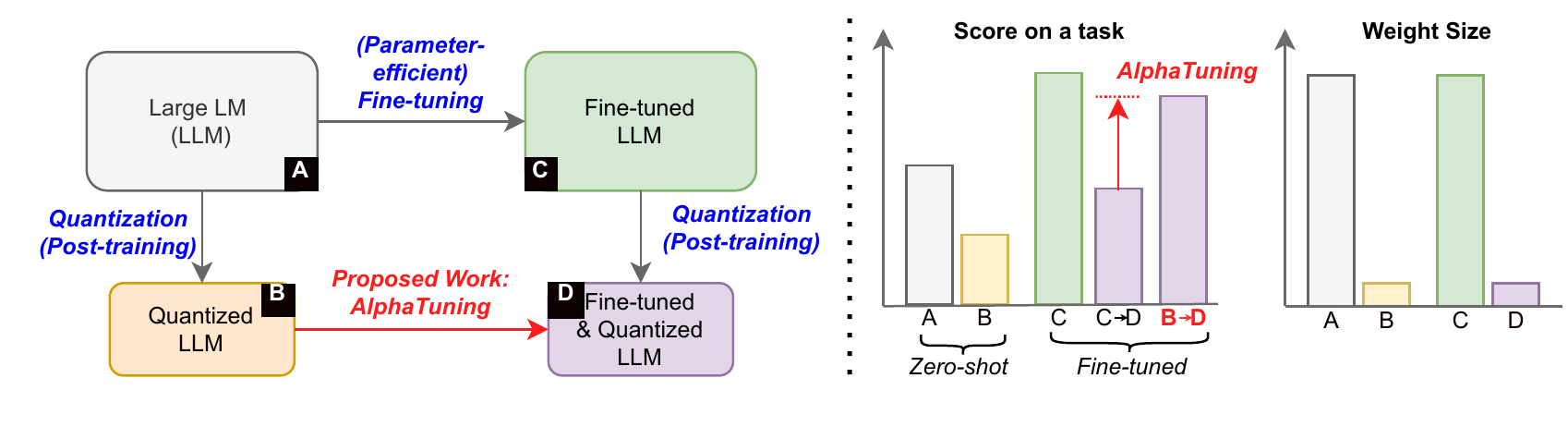}
	\caption{Approaches to satisfy both parameter-efficient adaptation and parameter quantization. Our proposed AlphaTuning technique can achieve 1) competitive performances to fine-tuned LMs (i.e., A$\leadsto$C) with a remarkably reduced parameter size, and 2) significantly better scores than quantized LMs implemented through A$\leadsto$C$\leadsto$D.}
	\label{fig:intro_overview}
\end{figure*}

Although trainable parameters can be significantly reduced by parameter-efficient adaptation schemes, we notice that the memory footprints for inference are not reduced compared to those of PLMs\footnote{In practice, the adaptation is usually implemented by adding small additional parameters to PLMs.}.
To enable efficient deployments of multiple downstream tasks, we incorporate model compression and parameter-efficient adaptation.
We argue that previous model compression techniques were not practical solutions in terms of parameter-efficiency for adaptations.
For example, Quantization-Aware Training (QAT)~\cite{jacob2018quantization, lsq} can perform full fine-tuning coupled with model compression; however, each task needs dedicated memory storage as much as that of a compressed PLM.
Our key observation to achieve a compression-aware parameter-efficient adaptation is that, once a PLM is quantized, only a small amount of quantization-related parameters is needed to be fine-tuned for each target task.
As a result, both the overall memory footprints and the number of trainable parameters for adaptation can be substantially reduced.



Figure~\ref{fig:intro_overview} illustratively compares two different approaches enabling both model compression and parameter-efficient adaptation.
Fine-tuned and quantized LMs can be achieved through A$\leadsto$C$\leadsto$D or A$\leadsto$B$\leadsto$D as shown in Figure~\ref{fig:intro_overview}.
In the case of A$\leadsto$C$\leadsto$D, we may have a large number of trainable parameters, and/or PTQ may degrade performance on downstream tasks.
To address such issues, we investigate A$\leadsto$B$\leadsto$D scheme, called ``AlphaTuning'' in this work.
Specifically, we factorize the parameters of large PLMs into binary values and scaling factors.
Then, AlphaTuning conducts the adaptation by training only the scaling factors that occupy a small portion in the quantization format, while freezing the other binary values.
Note that, to conduct A$\leadsto$B, we consider post-training quantization (PTQ) \cite{OCS, adaquant, brecq} because the QAT demands significant computational overhead for training from a scratch with the whole dataset.

In this paper, our contributions are as follows:
\begin{itemize}
    \item To the best of our knowledge, this work is the first successful compression-aware parameter-efficient adaptation method.
    \item We report that once PLMs are quantized by PTQ, training scaling factors (less than 0.1\% of total parameter size) for each task only is enough for successful adaptations.
    \item Throughout various LMs and tasks, we demonstrate that AlphaTuning can achieve high scores even under 4-bit quantization.
\end{itemize}

\section{Recent Work}

\paragraph{Large-Scale Language Models and Quantization}

Pre-trained transformer-based language models \cite{BERT,gpt2} have shaped the way we design and deploy NLP models. 
In recent years, the explosion of availability of large-scale (\ie, larger than ten-billion scale) language models \cite{gpt3,gpt-neo,palm,zhang2022opt,hoffmann2022training} has paved way for a new era in the NLP scene, where few-shot learning and the parameter-efficient adaptation for downstream tasks will be more important \cite{he2021towards}. 
The quantization (that we discuss in detail in the next section) is an effective approach to fundamentally overcome the space and time complexities of the large-scale language models \cite{zafrir2019q8bert,qualcommbertquant}, but existing methods are only applicable to limited domains and task adaptability under the quantized state.

\paragraph{Parameter-Efficient Adaptation of LMs}

Adapting language models efficiently for a task and domain-specific data has been at the center of the community's interests since the emergence of large-scale language models. 
One promising approach is in-context learning (ICL) \cite{gpt3}, in which the language model learns and predicts from the given prompt patterns. 
As the technique elicits reasonable few-shot performances from the large-scale language models without parameter-tuning, a plethora of works \cite{zhao2021calibrate,lu2022fantastically,reynolds2021prompt,min2022rethinking} have investigated the underlying mechanism and proposed various methods to further exploit this approach. 
Another class of techniques is to adopt external or partially internal parameters such as continuous prompt embeddings to enable parameter-efficient LM adaptation, which is based on the intuition that specific prompt prefixes may better elicit certain LM behaviors. Earlier works explored the discrete prompt token space \cite{shin2020autoprompt}, but later work showed that optimizing on the continuous word embedding space yielded better results \cite{p-tuning,prefixtuning,gu2022ppt}, even performing on par with full fine-tuning \cite{lester2021power,vu2022spot}. Another similar line of works explored introducing new parameters within the Transformer blocks or partially training existing parameters \cite{adapter,zhang2020beyond,karimi2021compacter,lora}. Finally, some works have suggested unifying all existing approaches related to parameter-efficient fine-tuning \cite{he2021towards,zhang2022hyperpelt}.

\begin{table*}[t]
    \centering
    \small
    \begin{tabular}{lcccccccc}
        \Xhline{2\arrayrulewidth}
        \mr{2}{Layer}   & $\mW$ Shape    & $\mW$ Size   & \mr{2}{$g$} & $\alpha \in \R$ & $B \in \{$-1,+1$\}$ & \multicolumn{3}{c}{Quantized $\mW$ Size (MB)} \\ \cline{7-9}
     & ($h_{out}$, $h_{in}$) & (FP32) & & Shape & Shape & $q=1$ & $q=2$ & $q=3$   \\ 
        \Xhline{2\arrayrulewidth}
        ATT\_qkv  & ($3h,h$) & 12.58 MB   & $h$ & ($q,3h$) & ($q,3h,h$) & 0.41 & 0.81 & 1.22   \\ \hline
        ATT\_output & ($h,h$)  & 4.19 MB   & $h$ & ($q,h$) & ($q,h,h$) & 0.14 & 0.27 & 0.41   \\ \hline
        FFN\_h\_4h & ($4h,h$)   & 16.78 MB  & $h$ & ($q,4h$) & ($q,4h,h$) & 0.54 & 1.08 & 1.62  \\ \hline
        \mr{3}{FFN\_4h\_h} & \mr{3}{($h,4h$)} & \mr{3}{16.78 MB}  & 4$h$ & ($q,h$) & ($q,h,4h$) & 0.52 & 1.06 & 1.56  \\  \cline{4-9}
         & & & $h$ & ($q,4h$) & ($q,4h,h$) & 0.54 & 1.08 & 1.62  \\\cline{4-9}
         & & & 0.5$h$ & ($q,8h$) & ($q,8h,h$) & 0.56 & 1.11 & 1.67  \\ 
        \Xhline{2\arrayrulewidth}
    \end{tabular} 
    \caption{BCQ scheme for $q$-bit quantization applied to linear layers of the Transformers and examples of BCQ formats for GPT-2 medium model (hidden size $h$ is 1024). Row-wise quantization is performed when $g=h_{in}$. Lower $g$ results in slightly increased weight size after quantization.}
    \label{table:quant_example}
\end{table*}

\section{Quantization for AlphaTuning}

Enterprise-scale LMs, such as 175B GPT-3, face challenges in the prohibitive cost of massive deployment mainly resulting from their huge parameter size.
To facilitate cost-effective LMs by alleviating memory requirements without noticeable performance degradation, we can consider compression techniques, such as quantization \citep{jacob2018quantization}, pruning \citep{frankle2020pruning}, and low-rank approximation \citep{SVD2013}.
Memory reduction by model compression is also useful to reduce latency because memory-bound operations dominate the overall performance of LMs with a small batch size \citep{sc22}.
In addition, model compression can save the number of GPUs for inference because GPUs present highly limited memory capacity \citep{megatron-1st, megatron-2nd}.
In this work, we choose quantization as a practical compression technique because of its high compression ratio, simple representation format, and the capability to accelerate memory-bound workloads \citep{Extremely_low_quant_emnlp2020}.

Let us discuss our quantization strategy for LMs (see more details in Appendix C).
We choose non-uniform quantization since uniform quantization demands aggressive activation quantization (to exploit integer arithmetic units) which is challenged by highly non-linear operations (such as softmax and layer normalization) of the Transformers \cite{qualcommbertquant}.
Even though uniform quantization can mitigate performance degradation by frequent activation quantization/dequantization procedures \cite{int8_transformer} or additional high-precision units \cite{i-bert}, such techniques are slow and/or expensive.
Among various non-uniform quantization formats, we choose binary-coding-quantization (BCQ) \cite{Greedy_Quant, xu2018alternating} which is extended from binary neural networks \cite{rastegariECCV16} because of high compression ratio \cite{Extremely_low_quant_emnlp2020} and efficient computations \cite{xu2018alternating, biqgemm}.

\begin{figure}[t]
    \centering
    \includegraphics[width=1.0\linewidth]{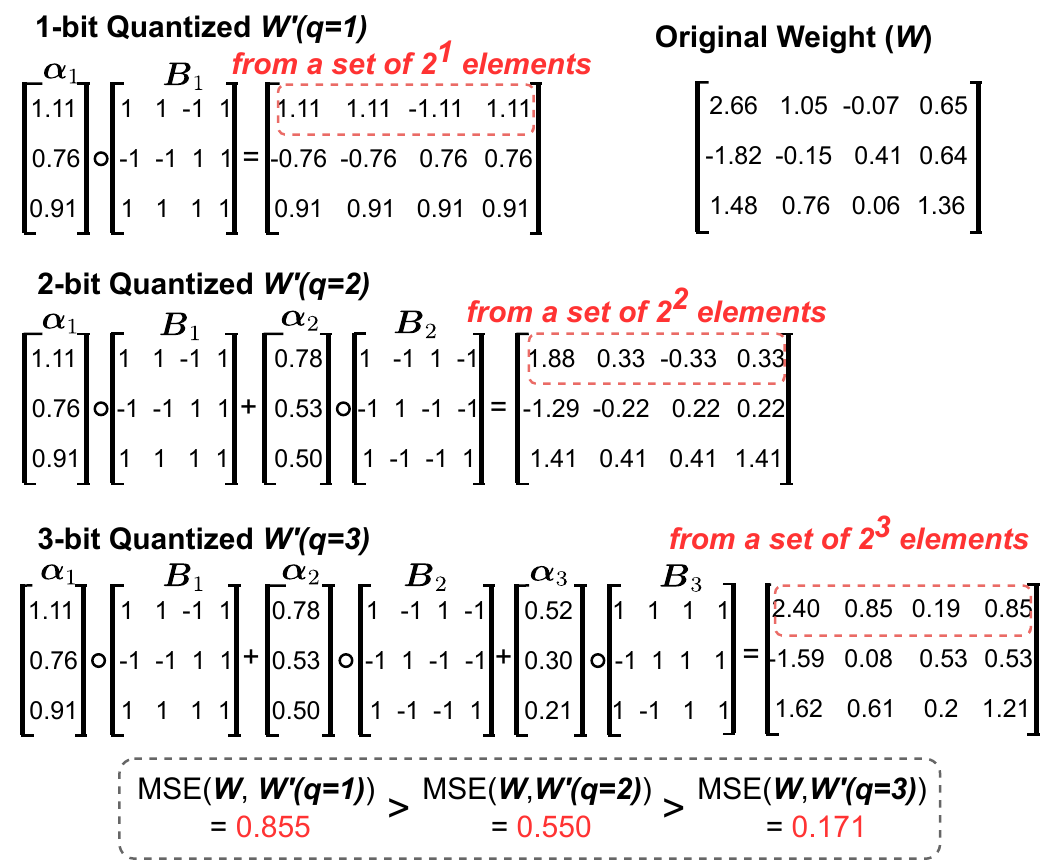}
	\caption{BCQ examples with $g=4$ and different $q$ values. As $q$ increases, the MSE between the original weight and the quantized weight decreases.}
	\label{fig:quant_example}
\end{figure}



\paragraph{BCQ Format}
Given a full-precision weight vector $\vw \in \R^g$, BCQ format approximates $\vw$ to be $\vw \approx \sum_{i=1}^{q} \alpha_i \vb_i$ where $q$ is the number of quantization bits, $\alpha \in \R$ is a scaling factor to be shared by $g$ weights, and $\vb \in \{-1,+1\}^g$ is a binary vector.
Note that $g$ represents a group size or the number of weights sharing a common scaling factor.
Thus, $g$ is a hyper-parameter for quantization.
When $q$=1, $\alpha$ and $\vb$ can be analytically determined to minimize the mean squared error (MSE).
If $q>1$, however, $\alpha$ and $\vb$ need to be obtained by heuristic methods such as greedy approximation \cite{Greedy_Quant} and iterative fine-tuning method \cite{xu2018alternating}.

\begin{figure*}[t]
    \centering
    \includegraphics[width=1.0\linewidth]{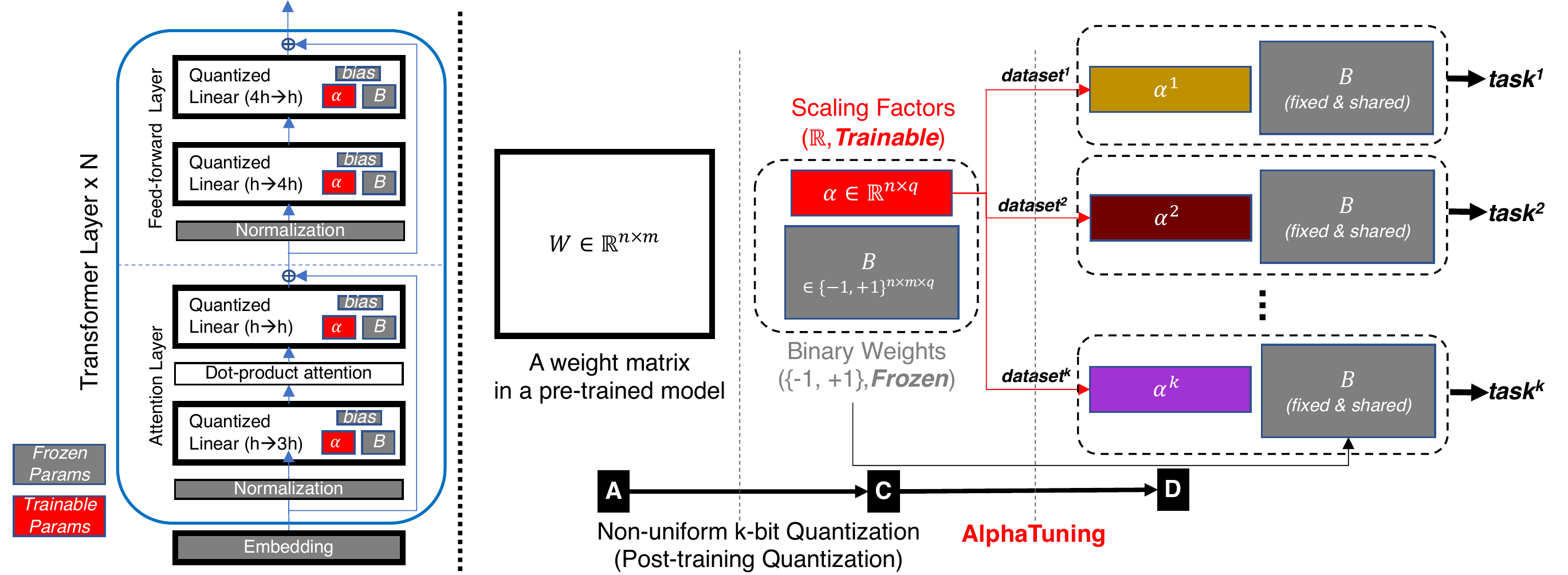}
    \caption{(Left): Quantized Transformer structure in which parameters are categorized into frozen ones and trainable ones. (Right): Overview of AlphaTuning process that trains scaling factors only for adaptation.}
	\label{fig:alphatuning_overview}
\end{figure*}

For a weight matrix $\mW\in\R^{h_{out}{\times}h_{in}}$, row-wise quantization (i.e., $g=h_{in}$) is a popular choice\footnote{On/off-chip memory bandwidth can be maximized by contiguous memory allocation if row-wise quantization is adopted. Additionally, for large LLMs (along with a large $h_{in}$), the amount of $\bm{\alpha}$ becomes almost ignorable (\ie, $\bm{\alpha}$ size is $32 / h_{in}$ of $\bm{\mB}$ size) even assuming 32 bits to represent an $\alpha$.} \cite{biqgemm, xu2018alternating} and can be expressed as follows:
\begin{equation}
    \mW \approx \sum_{i=1}^q \diag(\bm{\alpha}_i) \cdot \mB_i,
    \label{eq:1}
\end{equation}
where $\bm{\alpha}_i \in \R^{h_{out}}$, $\mB_i \in {\{-1, +1\}}^{h_{out}{\times}h_{in}}$, and $\text{diag}(\cdot)$ denotes the function of a vector that outputs a zero-matrix except for the vector elements in its diagonal.
A linear operation of $\mY=\mX\cdot(\mW)^\top$, then can be approximated as follows:
\begin{equation}
\begin{aligned}
     \mY &= \mX \cdot \mW^\top \\
         &\approx \mX \cdot \left(\sum_{i=1}^q \diag\left(\bm{\alpha}_i\right) \cdot \mB_i\right)^\top  \\
         &= \sum_{i=1}^q \Big((\mX \cdot \mB_i^\top) \cdot \diag\left(\bm{\alpha}_i\right)\Big),
\end{aligned}
\label{eq:2}
\end{equation}
where $\mX \in \R^{n_{b}{\times}h_{in}}$, and $\mY \in \R^{n_{b}{\times}h_{out}}$.
Note that even though $\mX$ is not quantized above, most complicated floating-point operations are removed due to binary values in $\mB$.
Since the computational advantages of BCQ have been introduced in the literature \citep{Hubara2016, biqgemm}, we do not quantize activations in this work to improve quantization quality.


Figure~\ref{fig:quant_example} describes the row-wise BCQ examples based on greedy approximation \cite{Greedy_Quant} when $q$ varies.
Note that increasing $q$ and/or decreasing $g$ can reduce the MSE after quantization at the cost of a lower compression ratio.

\paragraph{Transformer Quantization}
Table~\ref{table:quant_example} presents our BCQ scheme applied to linear layers of the Transformers while BCQ formats are illustrated for the medium-sized GPT-2 model (that has a hidden size ($h$) of 1024).
Note that if $g$ is large enough such that each scaling factor is shared by many weights, the amount of scaling factors is ignorable compared to that of $\mB$.
In Table~\ref{table:quant_example}, hence, 1-bit quantization attains almost 32$\times$ compression ratio compared to FP32 format while lower $g$ slightly increases storage overhead induced by additional scaling factors.



\section{AlphaTuning: Efficient Fine-Tuning of Quantized Models}



\begin{table*}[t]
\small
    \centering
    \begin{threeparttable}
    \begin{tabular}{clcccccccc}
        \Xhline{2\arrayrulewidth}
        \mr{2}{Model} & \mr{2}{Method} & \mr{2}{$q$}    & Trainable & \multicolumn{2}{c}{Model Size} & \mr{2}{Valid\\Loss} & \multicolumn{3}{c}{BLEU (95\% Confidence Interval)} \\  \cmidrule(lr){5-6}  \cmidrule(lr){8-10}
         & &                                            & Params    & CKPT & Total  & & Unseen & Seen & All \\
        \Xhline{2\arrayrulewidth}
        \mr{7}{GPT-2\\M} 	& FT (Fine-Tuning)  & - & 354.9M & 1420MB & 1420MB      & 0.79     & $32.7_{\pm.6}$ & $62.0_{\pm.4}$ & $48.4_{\pm.3}$ \\
                    & $\Rightarrow$ \PTQA \tnote{1}	        & 3 & - & 327MB & 327MB      & 2.03  & $25.0_{\pm2.5}$ & $58.7_{\pm1.0}$ & $43.2_{\pm3.3}$ \\
                    
                    \Cline{1pt}{2-10}
                    & LoRA	                    & - & 0.35M  & 1.4MB & 1420MB       & 0.81 	& $45.5_{\pm.4}$ & $64.3_{\pm.2}$ & $55.8_{\pm.3}$    \\ 
                    & $\Rightarrow$ \PTQB \tnote{2}           & 3 & -  & 1.4MB & 328MB  & 2.98 & $15.8_{\pm3.0}$ & $15.8_{\pm3.4}$ & $15.8_{\pm3.2}$      \\ 
                    & $\Rightarrow$ \PTQC \tnote{3}           & 3 & -  & 327MB & 327MB       & 3.36  & $12.6_{\pm4.1}$  & $16.6_{\pm6.7}$ & $13.6_{\pm7.5}$ \\
                    \Cline{1pt}{2-10}
                    & AlphaTuning               & 3 & 0.22M  & 0.9MB & 327MB      & 0.81  & $40.9_{\pm.5}$ & $63.2_{\pm.5}$ & $53.1_{\pm.4}$ \\ 
                    & AlphaTuning	            & 2 & 0.22M  & 0.9MB & 289MB       & 0.84  & $37.3_{\pm.5}$ & $62.6_{\pm.5}$ & $51.3_{\pm.5}$  \\
        \Xhline{2\arrayrulewidth}
        \mr{8}{GPT-2\\L} 	& FT (Fine-Tuning)  & - & 774.0M & 3096MB & 3096MB  & 0.81     & $23.8_{\pm.3}$ & $60.8_{\pm.1}$ & $43.0_{\pm.3}$  \\
                    & $\Rightarrow$ \PTQA	                    & 3 & - & 535MB & 535MB     & 1.90 & $23.2_{\pm.8}$ & $62.7_{\pm.2}$ & $43.7_{\pm.7}$ \\
                    \Cline{1pt}{2-10}
                    
                    & LoRA                      & - & 0.77M & 3.1MB & 3096MB   & 0.79 	& $48.4_{\pm.3}$ & $64.0_{\pm.3}$ & $57.0_{\pm.1}$     \\ 
                    & $\Rightarrow$ \PTQB                     & 3 & - & 3.1MB & 538MB & 1.97 & $20.1_{\pm5.2}$ & $27.8_{\pm4.1}$ & $24.1_{\pm4.5}$ \\
                    & $\Rightarrow$ \PTQC                     & 3	& - & 535MB & 535MB	    & 1.97 & $14.0_{\pm7.2}$ & $26.6_{\pm11.5}$ & $25.8_{\pm13.0}$ \\ 
                    \Cline{1pt}{2-10}
                    & AlphaTuning               & 3 & 0.42M & 1.7MB & 535MB   & 0.84  & $47.0_{\pm.6}$ & $62.2_{\pm.2}$ & $55.3_{\pm.3}$                \\   
                    & AlphaTuning               & 2 & 0.42M & 1.7MB & 445MB   & 0.82  & $42.7_{\pm.4}$ & $62.9_{\pm.4}$ & $53.8_{\pm.1}$                \\  
                    & AlphaTuning               & 1 & 0.42M & 1.7MB & 355MB   & 0.87  & $28.1_{\pm.3}$ & $62.3_{\pm.7}$ & $47.1_{\pm.4}$                 \\

        \Xhline{2\arrayrulewidth}
    \end{tabular}
    \begin{tablenotes}
        \item[1] Fully fine-tuned LMs are quantized by PTQ using the Alternating method. 
        \item[2] For inference, quantized PLMs (by the Alternating method) are dequantized to be merged with trainable parameters for LoRA. This method is parameter-efficient but we have low scores and dequantization overhead.
        \item[3] After LoRA, frozen weights and trainable weights are merged and then quantized (by the Alternating method). Since PTQ is applied to the merged weights, each task needs to store the entire (quantized) model.
  \end{tablenotes}
    \end{threeparttable}
    \caption{Validation loss and test scores on WebNLG with various adaptation methods using GPT-2 models (see Table~\ref{app:table:lrwd} in Appendix for  hyper-parameter selections and Table~\ref{app:table:WebNLG} in Appendix additional scores). For full fine-tuning and LoRA, we explored learning rates and weight decay factors while the other hyper-parameters are from \cite{lora}. $g$ is selected to be $h_{in}$ in each layer for row-wise quantization.}
    \label{table:WebNLG}
\end{table*}

\subsection{AlphaTuning Principles}


The key idea of AlphaTuning is identifying parameters presenting greater expressive power to minimize the number of trainable parameters after PTQ.
Note that training affine parameters (that transform the activations through operations such as scaling, shifting, and rotating) reportedly achieves reasonably high accuracy even when all the other parameters are fixed to be random \cite{frankle2020training}.
Interestingly, scaling factors obtained by the BCQ format can be regarded as affine parameters as shown in Eq.~\ref{eq:2}.
Based on such observation, Figure~\ref{fig:alphatuning_overview} presents the overview of AlphaTuning.
First, we quantize the weights of linear layers of the Transformers that dominate the overall memory footprint \cite{sc22}.
Then, the BCQ format factorizes the quantized weights into scaling factors and binary values.
Finally, the scaling factors are trained for a given target task and all the other parameters (\eg, biases, binary values $\mB$, and those of the normalization layer and embedding layer) are frozen regardless of downstream tasks. 


\paragraph{Training Algorithm}

For a linear layer quantized by Eq.~\ref{eq:1}, the forward propagation can be performed without dequantizing $\mW$ and be described as Eq.~\ref{eq:2}.
Similarly, the backward propagation can also be computed in the quantized format and the gradients of $\mW$ and $\bm{\alpha}$ with respect to $\mY$ (to conduct the chain rule) are obtained as follows:
\begin{equation}
    \partial {\mX} = \partial \mY \cdot \Big(\sum_{i=1}^q \diag(\bm{\alpha}_i) \cdot \mB_i \Big)
    \label{eq:3}
\end{equation}
\begin{equation}
    \partial {\bm{\alpha}}_i = \frac{(\partial \mY)^\top \mX \mB_i^\top \cdot \mathds{1}^\top }{g_L} \, (1\le i \le q),
    \label{eq:4}
\end{equation}
where $\mathds{1}$ is an $h_{out}$-long all-ones vector and $g_L$ is the group size of the layer $L$. 
Note that dividing by $g_L$ is empirically introduced in Eq.~\ref{eq:4} to prevent excessively large $\alpha$ updates and to enhance the stability of training.
Even if $g_L \ne h_{in}$ (\ie, other than row-wise quantization), we still can utilize the same equations by using tiling-based approaches \cite{biqgemm}.

\subsection{AlphaTuning for GPT-2}


We apply AlphaTuning to GPT-2 medium and large on WebNLG \cite{webnlg_2017} to explore a hyper-parameter space and investigate the effects of AlphaTuning as shown in Table~\ref{table:WebNLG}.
Note that in this paper, we assume that parameters (including $\alpha$) are represented as 32-bit floating-point numbers (\ie, FP32 format) unless indicated to be compressed by $q$-bit quantization.

\begin{table*}[t]
\small
    \centering
    \begin{tabular}{lcccccc}
        \Xhline{2\arrayrulewidth}
        Hyper-Parameter & Base & Trial & Loss & Unseen & Seen & All \\
        \Xhline{2\arrayrulewidth}
        
        Trainable Params   & 0.22M ($\bm{\alpha}_1$)   & 0.66M ($\bm{\alpha}_1$,$\bm{\alpha}_2$,$\bm{\alpha}_3$)     & 0.76 & $40.6\pm.4$ & $63.2\pm.2$ & $53.1\pm.1$  \\ 
        Dropout Rate                    & 0.0                           & 0.1                       & 0.81 & $42.4\pm.3$ & $61.2\pm.4$ & $52.7\pm.2$ \\
        PTQ Method         & Greedy                  & Alternating         & 0.80 & $41.0\pm.6$ & $63.0\pm.3$ & $53.0\pm.3$ \\
        LR Warm-up                      & 0 steps                       & 500 steps                 & 0.81 & $41.0\pm.2$ & $63.3\pm.1$ & $53.3\pm.1$ \\
        Epochs                      & 5 epochs                      & 3 epochs                  & 0.82 & $42.2\pm.6$ & $62.9\pm.4$ & $53.6\pm.4$ \\
        Epochs                      & 5 epochs                      & 10 epochs                 & 0.82 & $38.5\pm.7$ & $62.7\pm.5$ & $51.9\pm.4$ \\ \hline
        \multicolumn{3}{c}{Base Hyper-Parameter Selection (Table~\ref{table:WebNLG})}                            & 0.81  & $40.9\pm.5$ & $63.2\pm.5$ & $53.1\pm.4$ \\   
        \Xhline{2\arrayrulewidth}
    \end{tabular}
    \caption{Experimental results on WebNLG to investigate the impact of hyper-parameter selection for AlphaTuning on GPT-2 medium quantized by PTQ using 3-bit quantization (i.e., $q=3$). Test BLEU scores are averaged over 5 trials with the same learning rates and weight dacay fectors in Table~\ref{table:WebNLG}.}
    \label{table:WebNLG_ablation}
\end{table*}

\paragraph{Adaptation Details}

PTQ for AlphaTuning is performed on the pre-trained GPT-2 by the Greedy method \cite{Greedy_Quant}.
Then, for $q$-bit quantization, we train only $\bm{\alpha}_1$ among $\bm{\alpha}_1 \cdots \bm{\alpha}_q$  to maximize parameter-efficiency of adaptation because training all $\bm{\alpha}$ values provides only marginal gains as shown in Table~\ref{table:WebNLG_ablation}.
Training $\bm{\alpha}_1$ is performed by a linear decay learning rate schedule without dropout.
For each hyper-parameter selection, test scores are measured at the 5\textsuperscript{th} epoch and averaged over 5 trials (along with 5 random seeds which are fixed for the experiments in Table~\ref{table:WebNLG_ablation} justifying our hyper-parameter selections).
For all adaptation methods considered in Table~\ref{table:WebNLG}, learning rates and weight decay factors are explored to produce the best at `all' category (see Table~\ref{app:table:webnlg-lr} for exploration results on AlphaTuning).


\begin{figure}[t]
     \centering
     \begin{subfigure}[]{\linewidth}
         \centering
         \includegraphics[width=\linewidth]{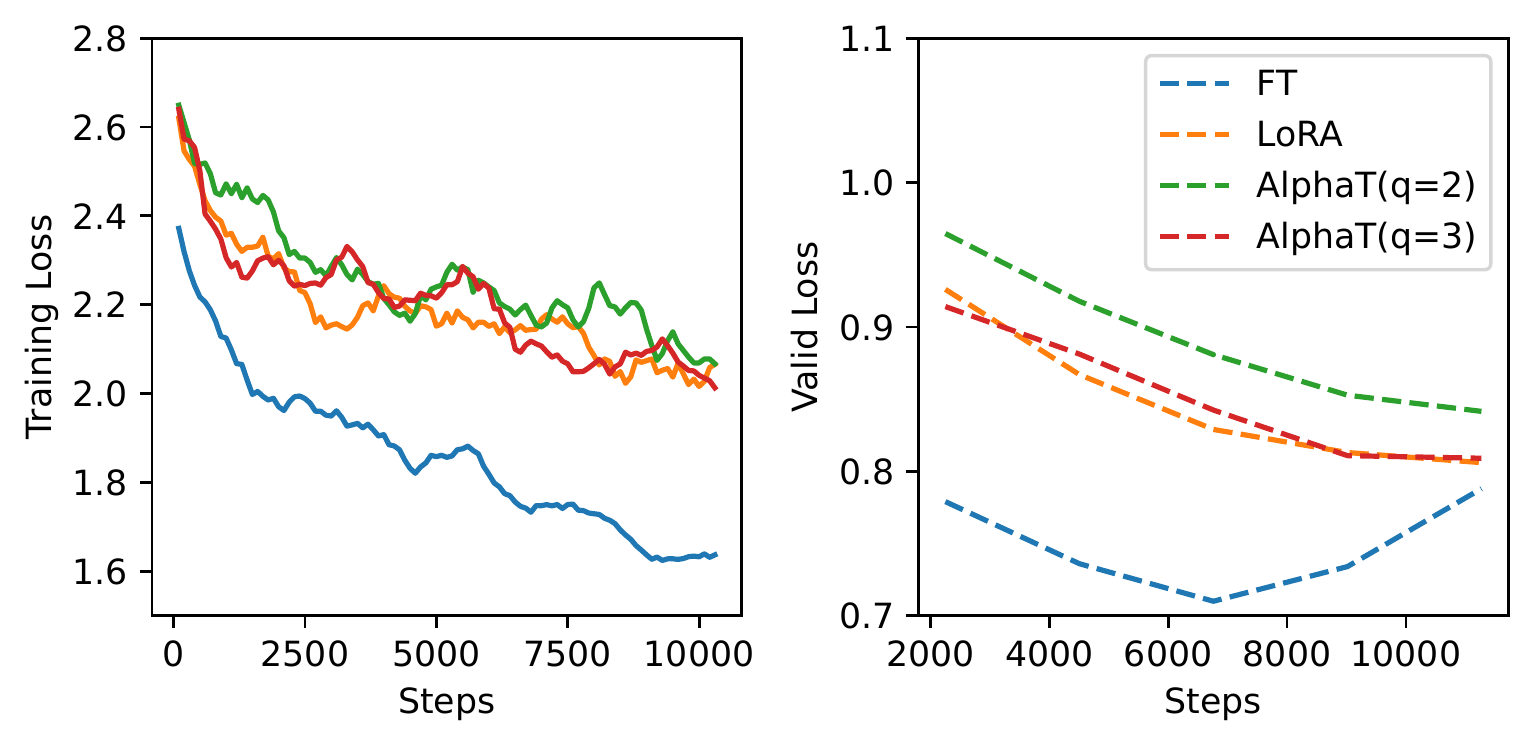}
         \caption{GPT-2 Medium}
         \label{fig:wennlg_graph_m}
     \end{subfigure}
     
     \begin{subfigure}[]{\linewidth}
         \centering
         \includegraphics[width=\linewidth]{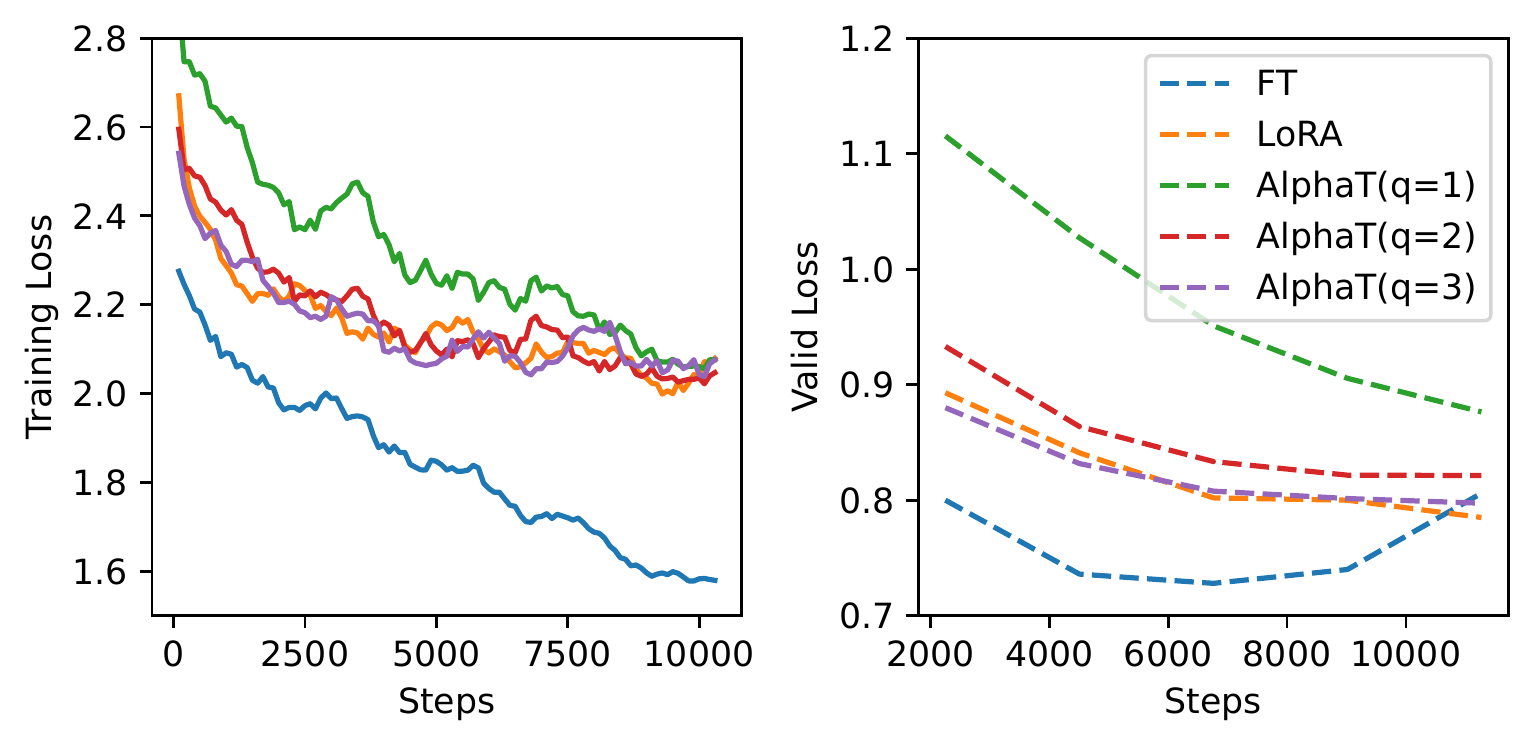}
         \caption{GPT-2 Large}
         \label{fig:wennlg_graph_l}
     \end{subfigure}
        \caption{Training/validation loss on WebNLG by full fine-tuning, LoRA, and AlphaTuning ($q=2$ or 3).}
        \label{fig:webnlg_graph}
\end{figure}

\paragraph{Comparison with Fine-Tuning and LoRA}

We compare AlphaTuning with full fine-tuning and LoRA reproduced by using hyper-parameters (except learning rates and weight decay factors) in \cite{lora} for WebNLG.
As shown in Table~\ref{table:WebNLG}, AlphaTuning provides BLUE scores which are comparable to that of LoRA and better than that of full fine-tuning, while both total memory footprint and checkpoint (CKPT)\footnote{Indicates dedicated storage for each downstream task.} memory sizes are significantly reduced.
The different scores can be partly explained by Figure~\ref{fig:webnlg_graph} showing that the training process by AlphaTuning or LoRA converges well while the full fine-tuning causes overfitting.
Interestingly, even though we train only $\bm{\alpha}_1$ (and hence, $\bm{\alpha}_2$ and $\bm{\alpha}_3$ are fixed for all tasks), increasing $q$ improves the validation loss and the test BLEU scores.
Note that as $q$ increases, `Unseen' scores are enhanced rapidly while `Seen' scores are not affected noticeably.
Overall, AlphaTuning with the 3-bit (\ie, $q=3$) quantization can be a successful parameter-efficient adaptation with a high compression ratio.



\begin{table}[t]
\small
    \centering
    \begin{tabular}{cccccc}
        \Xhline{2\arrayrulewidth}
        $g$     & Params    & Loss & Unseen     & Seen      & All   \\ \hline
        64 & 4.72M & 0.78 & $42.7_{\pm.3}$ & $64.2_{\pm.2}$ & $54.5_{\pm.2}$ \\
        256     &   1.18M   & 0.77	& $41.7_{\pm.4}$	& $63.9_{\pm.3}$	& $54.0_{\pm.2}$  \\
        512     &   0.59M   & 0.77	& $41.2_{\pm.7}$	& $63.7_{\pm.1}$	& $53.7_{\pm.3}$  \\
        1K     &   0.30M & 0.79 & $40.9_{\pm.6}$ & $63.6_{\pm.5}$ & $53.4_{\pm.3}$ \\
        $h_{in}$    &   0.22M   & 0.81	& $40.9_{\pm.5}$	& $63.2_{\pm.5}$	& $53.1_{\pm.4}$  \\
        2K    &   0.15M   & 0.84	& $40.9_{\pm.4}$	& $62.5_{\pm.7}$   & $52.8_{\pm.4}$  \\
        \Xhline{2\arrayrulewidth}
    \end{tabular}
    \caption{Impact of $g$ (group size) when AlphaTuning ($q$=3) is applied to GPT-2 medium on WebNLG. When $g=h_{in}$, row-wise quantization is indicated.}
    \label{table:WebNLG_groupsize}
\end{table}

\begin{table*}[t]
\small
    \centering
    \begin{tabular}{clcccccc}
        \Xhline{2\arrayrulewidth}
        Model & Method & $q$ & Trainable Params & Valid Loss & BLEU & METEOR & TER \\
        \Xhline{2\arrayrulewidth}
        \mr{4}{GPT-2\\Medium} 	& Fine-Tuning  & -   & 354.92M    & -     & $46.0_{\pm0.1}$ &	$0.39$ & $0.46$ \\
                    & LoRA  & -         & 0.35M	    & - 	& $47.1_{\pm0.2}$ & $0.39$ & $0.46$ \\ \Cline{1pt}{2-8}
                    & AlphaTuning & 3              & 0.22M      & 1.13  & $46.6_{\pm0.2}$ & $0.38$ & $0.48$  \\
                    & AlphaTuning & 2 	            & 0.22M 	    & 1.17  & $45.7_{\pm0.2}$ &	$0.38$ & $0.49$ \\
                    
        \Xhline{2\arrayrulewidth}
        \mr{4}{GPT-2\\Large} 	& FineTuning  & -       & 774.03M    & -     & $46.5_{\pm0.1}$ &	$0.39$    & $0.45$\\
                    & LoRA & - 	            & 0.77M	    & - 	& $47.5_{\pm0.2}$ & $0.38$ &	$0.45$\\ \Cline{1pt}{2-8}
                    & AlphaTuning & 3   & 0.42M      & 1.08 & $47.8_{\pm0.2}$ & $0.39$ & $0.47$  \\
                    & AlphaTuning & 2 & 0.42M      & 1.10 &	$47.2_{\pm0.2}$ &	$0.38$ &	$0.47$ \\
                    
        \Xhline{2\arrayrulewidth}
    \end{tabular}
    \caption{Test scores on DART with various adaptation methods using GPT-2 models (see  Table~\ref{app:table:lrwd} in Appendix for hyper-parameter selections). The checkpoint and weight sizes can be found in Table~\ref{table:WebNLG}. The results of full fine-tuning and LoRA are quoted from \cite{lora}. $g$ is selected to be $h_{in}$ in each layer for row-wise quantization. For all METEOR and TER scores in the table, the variances are less than 0.01.}
    \label{table:DART}
\end{table*}

\paragraph{Comparison with A$\leadsto$C$\leadsto$D in Figure~\ref{fig:intro_overview}}

As potentially alternative methods of AlphaTuning, we investigate the following three cases: 1) applying PTQ to a fully fine-tuned model (\ie, {\PTQA}), 2) applying PTQ to a PLM and then LoRA parameters are augmented (i.e., {\PTQB}), and 3) a PLM and LoRA parameters are merged and then quantized (\ie, {\PTQC}).
Such three cases induce various checkpoint sizes, total model sizes, and the number of trainable parameters as shown in Table~\ref{table:WebNLG}.
Note that the scores of {\PTQB} and {\PTQC} are degraded significantly.
In other words, model compression techniques and parameter-efficient adaptation methods may have conflicting properties when combined in a straightforward manner.
Even though {\PTQA} shows better scores than the other two cases, the number of trainable parameters remains to be the same as that of full fine-tuning and checkpoint size for a task is considerably larger than that of LoRA and AlphaTuning.
By contrast, AlphaTuning offers acceptable BLEU scores even with a smaller number of trainable parameters and a smaller checkpoint size than those three cases.



\paragraph{Hyper-Parameter Selection}

A few hyper-parameters (such as dropout rate and the number of epochs) are related to the trade-off between `Unseen' score and `Seen' score as described in Table~\ref{table:WebNLG_ablation}.
In the case of PTQ method, even when the Alternating method \cite{xu2018alternating} is employed with many iterations to further reduce MSE, the scores become similar to that of the Greedy method after adaptation.
As such, we choose the Greedy method for all tasks in this paper.
The learning rate warm-up seems to present random effects depending on PLM, downstream task, and $q$ selection.
The group size $g$ provides the clear trade-off between the trainable parameter size and test scores as shown in Table~\ref{table:WebNLG_groupsize}.
Unless stated otherwise, we choose $g=h_{in}$ (\ie, row-wise quantization) in this paper.




\section{Experimental Results}

\begin{table*}[t]
\small
    \centering
    \begin{tabular}{clcccccccc}
        \Xhline{2\arrayrulewidth}
        \mr{2}{Model} & \mr{2}{Method} & \mr{2}{$q$} & Trainable & Valid & \multicolumn{5}{c}{Test Scores} \\
                      &               &             &  Params & Loss & BLEU & NIST & METEOR & ROUGE\_L & CIDEr \\
        \Xhline{2\arrayrulewidth}
        \mr{8}{GPT-2\\M}
                    & Fine-Tuning	        & - & 354.92M   & 1.28	& $67.5_{\pm.2}$ &	$8.60_{\pm.02}$ & $46.4_{\pm.2}$ & $70.8_{\pm.2}$    & $2.40_{\pm.01}$ \\
                    & $\Rightarrow$ \PTQA 	& 3 &       -   & 1.19 & $67.5_{\pm.5}$ & $8.58_{\pm.04}$ & $46.3_{\pm.5}$ & $70.3_{\pm.1}$  & $2.39_{\pm.01}$ \\
                    \Cline{1pt}{2-10}
                    & LoRA	        & - & 0.35M	    & 1.16	& $70.2_{\pm.2}$ &	$8.80_{\pm.04}$ & $46.8_{\pm.1}$ & $71.7_{\pm.4}$   & $2.53_{\pm.01}$ \\ 
                    & $\Rightarrow$ \PTQB	    & 3 &      -	    & 4.10 & $11.1_{\pm5.3}$ & $2.35_{\pm1.04}$ & $12.4_{\pm4.2}$ & $29.9_{\pm7.8}$ & $0.35_{\pm.18}$\\ 
                    & $\Rightarrow$ \PTQC	    & 3 &      -	    & 4.38 & $7.5_{\pm3.2}$ & $1.31_{\pm.21}$ & $11.4_{\pm.8}$ & $29.8_{\pm3.0}$ & $0.29_{\pm.04}$\\ 
                    \Cline{1pt}{2-10}
                    & AlphaTuning    & 3 &	           0.22M 	    & 1.18	& $69.9_{\pm.3}$ & $8.79_{\pm.05}$  & $46.7_{\pm.2}$  & $71.7_{\pm.3}$ 	 & $2.51_{\pm.01}$  \\
                    & AlphaTuning    & 2 &        0.22M	    & 1.20	& $70.0_{\pm.4}$ & $8.80_{\pm.05}$ & $46.7_{\pm.1}$  & $71.6_{\pm.5}$ 	 & $2.51_{\pm.01}$  \\
        \Xhline{2\arrayrulewidth}
        \mr{8}{GPT-2\\L}
                    & Fine-Tuning	   & -     & 774.03M    & 1.31 & $67.2_{\pm.3}$ & $8.61_{\pm.05}$ & $46.3_{\pm.1}$ & $70.5_{\pm.3}$ & $2.37_{\pm.01}$ \\
                    & $\Rightarrow$  \PTQA  & 3          & - & 0.98 & $66.5_{\pm1.0}$ & $8.45_{\pm.10}$ & $45.7_{\pm.3}$ & $70.3_{\pm.5}$ & $2.37_{\pm.03}$ \\
                    \Cline{1pt}{2-10}
                    & LoRA	 & -           & 0.77M      & 1.13 & $69.8_{\pm.2}$ & $8.80_{\pm.03}$ & $46.6_{\pm.1}$ & $71.7_{\pm.1}$ & $2.51_{\pm.01}$ \\ 
                    & $\Rightarrow$ \PTQB	 & 3            & -	    &1.87 & $50.9_{\pm3.4}$ & $6.63_{\pm.32}$ & $38.7_{\pm1.9}$ & $60.6_{\pm1.9}$ & $1.30_{\pm.19}$ \\ 
                     & $\Rightarrow$ \PTQC & 3 	            & -	    & 1.76 & $53.7_{\pm3.1}$ & $7.12_{\pm.37}$ & $40.0_{\pm1.5}$ & $61.7_{\pm1.1}$ & $1.5_{\pm.11}$\\ 
                     \Cline{1pt}{2-10}
                    & AlphaTuning & 2 	            & 0.42M	    & 1.14 & $69.7_{\pm.6}$ & $8.78_{\pm.08}$ & $46.6_{\pm.2}$ & $71.5_{\pm.3}$ & $2.51_{\pm.03}$ \\ 
                    & AlphaTuning & 1    	    & 0.42M	    & 1.18 & $69.7_{\pm.3}$ & $8.79_{\pm.03}$ & $46.6_{\pm.1}$ & $71.6_{\pm.2}$ & $2.51_{\pm.02}$ \\ 
                    \Xhline{2\arrayrulewidth}
    \end{tabular}
    \caption{Validation loss and test scores on E2E with various adaptation methods using GPT-2 models (see Table~\ref{app:table:lrwd} in Appendix for hyper-parameter selections). The number of trainable parameters, checkpoint sizes, and weight sizes are the same as in Table~\ref{table:WebNLG}. For full fine-tuning and LoRA, we explored learning rates and weight decay factors while the other hyper-parameters are quoted from \cite{lora}. $g$ is selected to be $h_{in}$ in each layer for row-wise quantization.}
    \label{table:E2E}
\end{table*}

\begin{table*}[t]
\small
    \centering
    \begin{tabular}{lcccccccc}
        \Xhline{2\arrayrulewidth}
        \mr{3}{Method}  & \mr{3}{$q$}  & \mr{3}{$g$}   & \multicolumn{3}{c}{MNLI} & \multicolumn{3}{c}{SAMSum} \\ \cmidrule(lr){4-6} \cmidrule(lr){7-9}
                        &              &               &  Trainable & Wight & Accuracy                  & Trainable & Weight & \mr{2}{R1 / R2 / RL} \\
                        &              &               & Params & Size  & (\%)                  & Params & Size   & \\
        \Xhline{2\arrayrulewidth}
        
                    Fine-Tuning  & - & -    & 1315.76M   & 5.26GB    & $83.6$                        & 1315.75M & 5.26GB  &	$49.4$ / $25.3$ / $40.5$ \\
                    $\Rightarrow$ \PTQA & 4         & $h$ & -  & 1.04GB    &     $76.7$           & - &   1.04GB  &	$13.0$ / $5.5$ / $11.4$ \\ \hline
                    AlphaTuning & 4 & $0.5h$ & 1.19M & 1.05GB &  $82.7$                            & 1.18M & 1.05GB    & $47.5$ / $24.1$ / $38.9$ \\
                    AlphaTuning & 4 & $h$ & 0.60M & 1.04GB &  $82.3$                                & 0.59M &  1.04GB    & $47.3$ / $23.2$ / $38.4$ \\
                    AlphaTuning & 3 & $0.5h$ & 1.19M  & 0.90GB & $82.4$                             & 1.18M & 0.90GB     & $47.4$ / $24.2$ / $39.0$ \\
                    AlphaTuning & 3 & $h$ & 0.60M  & 0.89GB & $82.4$                                 & 0.59M & 0.89GB    & $46.5$ / $22.8$ / $37.8$ \\

        \Xhline{2\arrayrulewidth}
    \end{tabular}
    \caption{Validation scores on MNLI dataset and test scores on SAMSum dataset with full fine-tuning and AlphaTuning using OPT 1.3B model for which hidden size $h$ is 2048 (see Appendix \ref{sec:app:opt_exp} for experimental details).}
    \label{table:OPT}
\end{table*}

To extensively demonstrate the influence of AlphaTuning, we apply detailed adaptation techniques and hyper-parameter selections that we explored by using GPT-2 models on WebNLG (in the previous section) to additional downstream tasks and OPT models \cite{zhang2022opt}.

\subsection{GPT-2 Models on DART and E2E}

Adaptations using full fine-tuning, LoRA, and AlphaTuning methods based on pre-trained GPT-2 medium/large are performed on DART \cite{nan2021dart} and E2E \cite{e2edata}.
As for DART dataset, we observe (in Table~\ref{table:DART}) AlphaTuning even with an extreme quantization (\eg, $q=2$) can maintain test scores to be similar to those of LoRA and full fine-tuning, both of which do not consider model compression.
In the case of E2E dataset, we find that 1) full fine-tuning suffers from degraded test scores, 2) even AlphaTuning with $q=1$ is a reasonable choice for GPT-2 large, and 3) quantizing a model (after being adapted by LoRA) destroys test scores.
All in all, when combined with pre-trained GPT-2 medium/large on various tasks, AlphaTuning turns out to be effective for both a high compression ratio and a massive reduction in the number of trainable parameters.

\subsection{OPT Models on MNLI and SAMSum}
We utilize a pre-trained OPT 1.3B model to be adapted through full fine-tuning or AlphaTuning on GLUE-MNLI \cite{williams-etal-2018-broad} and SAMSum \cite{gliwa2019samsum}.
For text classification on MNLI, an LM head layer is added on top of GPT-2 with randomly initialized weights \cite{gpt2}.
As evidenced by Table~\ref{table:OPT}, we find the following results: 1) {\PTQA} sometimes results in severely impaired scores (\eg, on SAMSum dataset) even when computations for PTQ are associated with a lot of iterations; 2) AlphaTuning outperforms {\PTQA} scheme (for the whole tasks in this paper), and 3) decreasing $g$ of AlphaTuning can improve scores.

\section{Discussion}

\paragraph{Memory during Adaptation}

As a compression-aware parameter-efficient adaptation technique, AlphaTuning reduces not only inference memory footprints (by quantization) and also training memory footprints during adaptation.
Specifically, optimizer states to be stored in GPU memory are derived only by scaling factors that occupy less than 0.1\% of total weight size if $g$ is large enough.
Such reduced GPU memory requirements during training correspond to increased batch size or a reduced minimum number of GPUs performing adaptation.

\paragraph{Embedding Layers}

In this work, we considered linear layers of the Transformers to be quantized by BCQ while embedding layers remain to be of full precision.
The rationale behind this choice is that as the model scales with a larger hidden size ($h$), the relative size of embedding layers becomes smaller.
To be more specific, the space complexities of linear layers and embedding layers follow $\mathcal{O}(h^2)$ and $\mathcal{O}(h)$, respectively.
As such, for large-scale LMs, we expect quantizing embedding layers to produce only marginal improvements on a compression ratio while test scores might be degraded.

\paragraph{Inference Speed}

As described in Eq.~\ref{eq:2}, BCQ format enables unique computations for matrix multiplications even when activations are not quantized.
Recent works \cite{biqgemm, sc22} show that matrix multiplications based on BCQ format can be expedited by the following operations: 1) compute all possible computations (combining partial activations and $\mB$) in advance and store them in look-up tables (LUTs) and 2) let LUT retrievals (using $\mB$ values as indices) replace floating-point additions in Eq.~\ref{eq:2}.
The major reasons for fast computations are due to byte-level accesses of LUTs and increased LUT reuse by increased $h$ \cite{biqgemm, sc22}.
Such LUT-based matrix multiplications can lead to latency improvement as much as a memory reduction ratio.

\section{Conclusion}

In this paper, we proposed AlphaTuning as the first successful compression-aware parameter-efficient adaptation method for large-scale LMs.
Through a few representative generative LMs (such as GPT-2), we demonstrated that once linear layers are quantized by BCQ format, training only scaling factors can obtain reasonably high scores.
We also empirically proved that quantizing an already adapted LMs would degrade scores significantly.
Incorporating various model compression techniques and parameter-efficient adaptation methods would be an interesting research topic in the future.

\section*{Limitations}
We believe that the major contributions in this paper would become more convincing as the size of the PLM increases, whereas the models used for experiments in this paper (\ie, GPT-2 and 1.3B OPT) may not be large enough compared to the large-scale LMs recently announced (\eg, OPT 175B).
Considering a few reports that larger models tend to be compressed by a higher compression ratio along with less performance degradation \cite{li2020train}, we expect AlphaTuning to be effective as well even for larger models, to say, of more than 10 billion of parameters.

The performance of AlphaTuning on 1.3B OPT becomes better than that of {\PTQA}, but inferior to that of the full fine-tuning.
We suspect such results might result from insufficient search of an appropriate training recipe for AlphaTuning.
Correspondingly, exploring learning hyper-parameters of AlphaTuning using larger LMs and more datasets would be required to yield general claims on the characteristics of AlphaTuning.

\section*{Ethics Statement}
Large language models (LMs) such as GPT-3~\citep{gpt3}, Gopher~\citep{gopher}, PaLM~\citep{palm}, and HyperCLOVA~\citep{hyperclova} have shown surprising capabilities and performances for natural language understanding and generation, in particular, an in-context zero or few-shot manner. They can provide innovative applications such as code generation~\citep{chen2021evaluating} and text-to-image generation~\cite{ramesh2021zero} via fine-tuning with additional data dedicated to each task. Despite their astonishing advantages, it is well known that large LMs have severe and challenging limitations for deployment to user applications, such as biased and toxic expression, hallucination, too heavy energy consumption, and carbon emission~\citep{weidinger2021ethical}. 
Our work aims to address the energy issue of large LMs in terms of inference and deployment. We expect that our method can alleviate energy consumption by reducing practical memory footprints and latency when the large LMs are deployed to various user applications. We might need to address the other ethical issues through further research for safer and better contributions of the large LMs. 



\bibliography{anthology,custom}
\bibliographystyle{acl_natbib}

\clearpage
\appendix

\begin{table*}[t]
\small
    \centering
    \begin{tabular}{clcccccccccc}
        \Xhline{2\arrayrulewidth}
        \mr{2}{Size} & \mr{2}{Method} & \mr{2}{$q$} & \multicolumn{3}{c}{BLEU} & \multicolumn{3}{c}{METEOR} & \multicolumn{3}{c}{TER} \\  \cmidrule(lr){4-6}  \cmidrule(lr){7-9}  \cmidrule(lr){10-12}
                     &                &             & U & S & A                 & U & S & A                   & U & S & A \\
        \Xhline{2\arrayrulewidth}
        \mr{7}{GPT\\M} 	
                    & FT(Fine-Tuning)  & - & $32.7_{\pm.6}$ & $62.0_{\pm.4}$ & $48.4_{\pm.3}$         & $.32$  &  $.45$  & $.39$ & $.63$ & $.33$ & $.47$ \\
                    & $\Rightarrow$ \PTQA     & 3        & $25.0_{\pm2.5}$  & $58.7_{\pm1.0}$  & $43.2_{\pm3.3}$    & $.28$  &  $.43$  &  $.36 $ &  $.87 $ &  $.37 $ &  $.60 $ \\
                    
                    \Cline{1pt}{2-12}
                    & LoRA	          & -  & $45.5_{\pm.4}$  & $64.3_{\pm.2}$  & $55.8_{\pm.3}$       & $.38$  &  $.45$  & $.42$ & $.47$ & $.32$ & $.39$ \\ 
                    & $\Rightarrow$ \PTQB     & 3        & $15.8_{\pm3.0}$ & $15.8_{\pm3.4}$  & $15.8_{\pm3.2}$     & $.20$  & $.21$  & $.21 $ & $1.03 $ &  $1.20 $ &  $1.12 $ \\
                    & $\Rightarrow$ \PTQC     & 3         & $12.6_{\pm4.1}$ & $16.6_{\pm6.7}$  & $13.6_{\pm7.5}$     & $.17$  & $.18$  & $.18 $ & $.72 $ & $.70 $ & $.68 $\\
                    \Cline{1pt}{2-12}
                    & AlphaTuning & 3      & $40.9_{\pm.5}$ & $63.2_{\pm.5}$  & $53.1_{\pm.4}$        & $.35$ &  $.44$  &  $.40 $ &  $.51 $ &  $.33 $ &  $.42 $ \\ 
                    & AlphaTuning & 2	   & $37.3_{\pm.5}$  & $62.6_{\pm.5}$ & $51.3_{\pm.5}$        & $.33$ &  $.44$  &  $.39 $ &  $.55 $ &  $.33 $ &  $.43 $ \\
                    
        \Xhline{2\arrayrulewidth}
        \mr{7}{GPT\\L} 	
                    & FT(Fine-Tuning)   & - & $23.8_{\pm.3}$ &  $60.8_{\pm.1} $ &  $43.0_{\pm.3} $ &  $.27 $ &  $.45 $ &  $.36 $ &  $.77 $ &  $.34 $ &  $.54 $\\
                    & $\Rightarrow$ \PTQA     & 3  & $ 23.2_{\pm.8}$ &  $62.7_{\pm.2} $ &  $43.7_{\pm.7} $ & $.27$ & $.45$  &  $.36$  &  $.77$  &  $.33$  &  $.54$  \\
                    \Cline{1pt}{2-12}
                    & LoRA	    & - & $48.4_{\pm.3} $ &  $64.0_{\pm.3} $ &  $57.0_{\pm.1}$ &  $.39 $ &  $.45 $ &  $.42 $ &  $.45 $ &  $.32 $ &  $.38 $\\ 
                    & $\Rightarrow$ \PTQB     & 3 &  $20.1_{\pm5.2}$ &  $27.8_{\pm4.1}$ &  $24.1_{\pm4.5} $ &  $.21$  &  $.25$ &  $.23$  &  $.99$  &  $.84$  &  $.91$  \\
                    & $\Rightarrow$ \PTQC	     & 3       &  $14.0_{\pm7.2}$&  $26.6_{\pm11.5}$ &  $25.8_{\pm13.0} $ & $.16$  & $ .23$  & $ .23$  & $ 1.24$  & $.76$  & $.79$ \\ 
                    \Cline{1pt}{2-12}
                    & AlphaTuning  & 3  &  $47.0_{\pm.6} $ &  $62.2_{\pm.2} $ &  $55.3_{\pm.3} $ &  $.38$  &  $.43$  &  $.41$  & $.46$  &  $.33 $ & $.39$  \\   
                    & AlphaTuning  & 2 &  $42.7_{\pm.4} $ &  $62.9_{\pm.4} $ &  $53.8_{\pm.1} $ &  $.36$  &  $.44$  &  $.40$  &	 $.49$  &  $.33 $ & $.41$ \\  
                    & AlphaTuning  & 1  &  $28.1_{\pm.3} $ &  $62.3_{\pm.7} $ &  $47.1_{\pm.4} $ &  $.29$  &  $.44$  &  $.36$  &  $.66$  &  $.33 $ & $.49$ \\
        \Xhline{2\arrayrulewidth}
    \end{tabular}
    \caption{Additional scores on WebNLG using GPT-2 (Extended results of Table~\ref{table:WebNLG}) including METEOR and TER scores (U: Unseen, S:Seen, A: All). Lower TER score indicates better generation capability while other scores are to be higher for better capability. For METEOR and TAR scores, the variances of all the cases are less than 0.01.}
    \label{app:table:WebNLG}
\end{table*}

\begin{figure}
     \centering
     \begin{subfigure}[]{\linewidth}
         \centering
         \includegraphics[width=\linewidth]{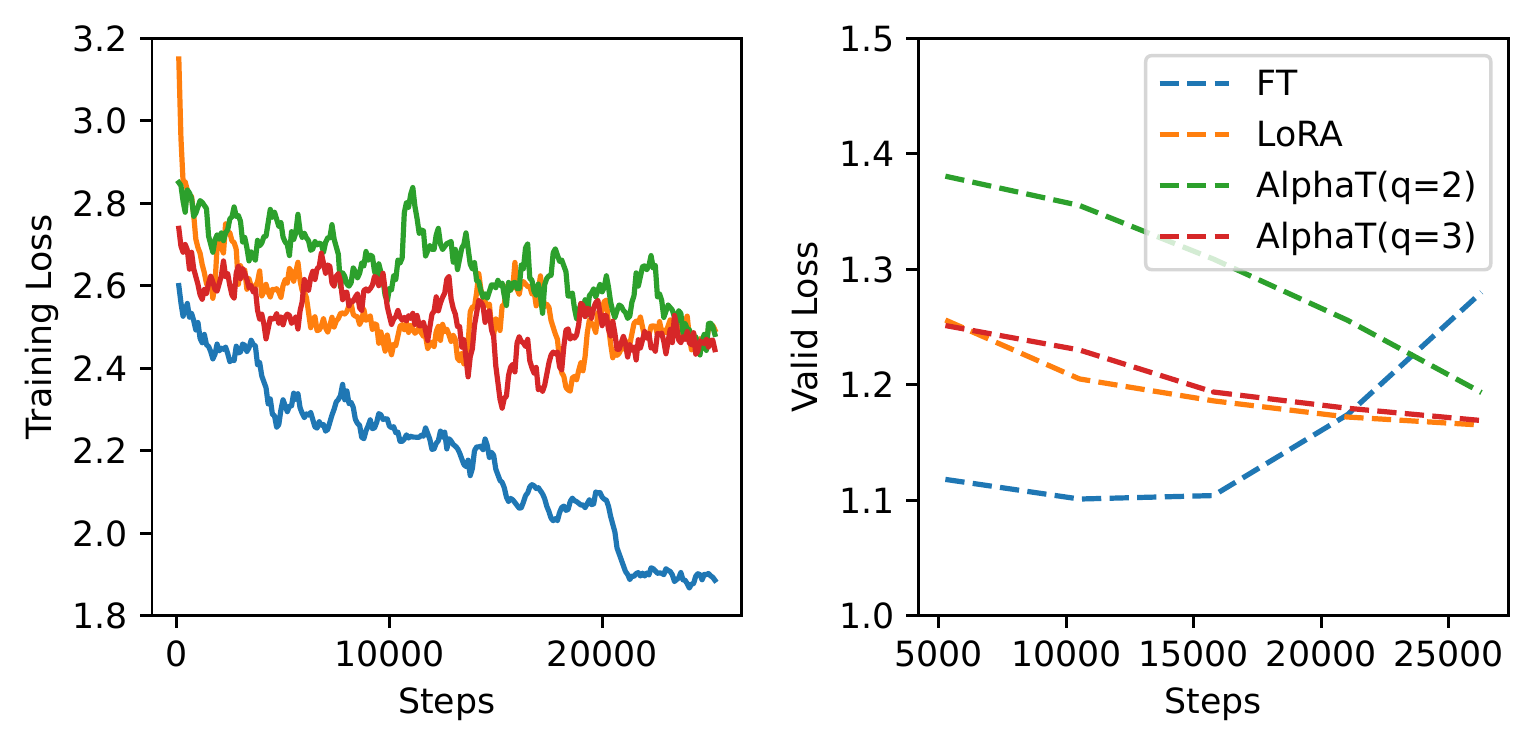}
         \caption{GPT-2 Medium}
         \label{fig:e2e_graph_m}
     \end{subfigure}
     
     \begin{subfigure}[]{\linewidth}
         \centering
         \includegraphics[width=\linewidth]{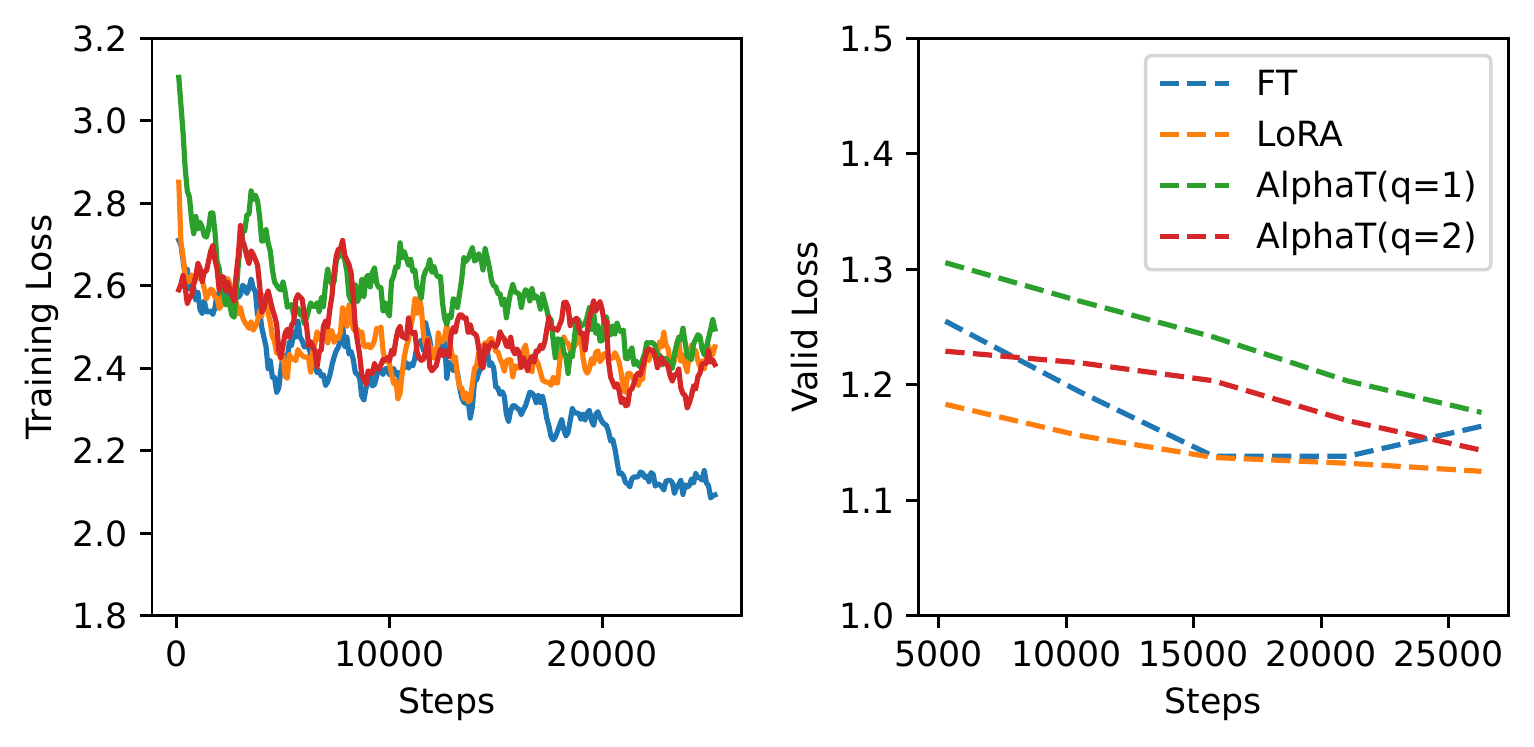}
         \caption{GPT-2 Large}
         \label{fig:e2e_graph_l}
     \end{subfigure}
        \caption{Training/validation loss on E2E by full fine-tuning, LoRA, and AlphaTuning ($q$=1, 2, or 3)}
        \label{fig:e2e_graph}
\end{figure}

\section{Experimental Details on GPT-2 Models}
\label{sec:app:exp_detail}
For all the adaptation experiments, we utilize the pre-trained GPT-2 Medium\footnote{Available at \url{{https://s3.amazonaws.com/models.huggingface.co/bert/gpt2-medium-pytorch_model.bin}}.}/Large\footnote{Available at \url{{https://s3.amazonaws.com/models.huggingface.co/bert/gpt2-large-pytorch_model.bin}}.}
models provided by \texttt{HuggingFace} \cite{huggingface}.
GPT-2 Medium consists of 24 Transformer layers with a hidden size ($h$) of 1024 and GPT-2 Large is composed of 36 Transformer layers with a hidden size of 1280.
Table~\ref{table:quant_example} includes the types of sublayers embedded into a GPT-2 layer.

\subsection{Dataset}
\textbf{WebNLG Challenge 2017}\cite{webnlg_2017} consists of 25,298 (data,text) pairs with 14 categories, which can be divided into nine ``Seen'' categories and five ``Unseen'' categories.
The model takes Resource Description Framework (RDF) triples as inputs and generates natural text descriptions to perform data-to-text generation task. 
Since the gradients during adaptation processes are calculated with only the ``Seen'' categories, the measured scores from ``Unseen'' categories are important for evaluating the generation performance of models.
In this paper, we represent three types of scores, `Unseen'(U), `Seen'(S), and `All'(A). 
Hyper-parameters are selected according to the best 'All' score.

\textbf{DART}(DAta Record to Text, \citet{nan2021dart}) is an open-domain text generation dataset with 82k examples, which are extracted from several datasets including WebNLG 2017 and Cleaned E2E.

\textbf{E2E} \cite{e2edata} was proposed for training end-to-end and data-driven natural language generation task. It consists of about 50k instances that provide meaning representations (MRs) and references for inference in the restaurant domain. Language models should perform data-to-text generation using suggested MRs.

\subsection{Adaptation Details}
For all the reproduced experiments and AlphaTuning experiments, AdamW \cite{adamw} optimizer and linear-decaying learning rate scheduler were used.
The number of epochs for the adaptation process is fixed to be 5 epochs and the other hyper-parameters are the same as reported in \citet{prefixtuning, lora}. 
We did not try to find the best results by evaluating and comparing the checkpoint at every epoch or by adjusting the number of epochs.
Instead, we explore the best results under varied learning rates and weight decay based on the reported list of hyper-parameters in \citet{prefixtuning} and \citet{lora} (the readers are referred to Table~\ref{app:table:lrwd}).

To evaluate the performance of the GPT-2 models, we use the beam search algorithm with several hyper-parameters listed in Table~\ref{table:app:decoding}.

\begin{table}[t]
\small
\centering
\begin{tabular}{lcc}
  \Xhline{2\arrayrulewidth}
              & FT/LoRA & AlphaTuning \\
               \Xhline{2\arrayrulewidth}
    Beam size & 10 & 10 \\
    Batch size & 16 & 16 \\
    No repeat ngram size & 4 & 4 \\
    Length penalty & 0.9 & 0.8 \\
     \Xhline{2\arrayrulewidth}
    
\end{tabular}
\caption{Hyper-parameters for beam search decoding}
\label{table:app:decoding}
\end{table}

\begin{table*}[t]
\small
    \centering
    \begin{tabular}{cclcccc}
    \Xhline{2\arrayrulewidth}
    \mr{2}{Dataset} & \mr{2}{Model} & \mr{2}{Method} & \multicolumn{2}{c}{Learning rate} & \multicolumn{2}{c}{Weight decay} \\ \cmidrule(lr){4-7}
     & & & best & range & best & range \\
    \Xhline{2\arrayrulewidth}
    \mr{9}{WebNLG \\(Table~\ref{table:WebNLG})}
            & \mr{4}{GPT\\M} 
                & FT & 1e-4 & \mr{2}{\{1e-4, 2e-4, 5e-4\}} & 0.01 & \mr{2}{\{0.0, 0.01, \\ 0.02\}} \\
            &   & LoRA & 5e-4 & & 0.01 &\\ \cline{3-7}
            &   & AlphaTuning($q{=}3$) & 1e-3 & \mr{2}{\{1e-4, 2e-4, 5e-4,\\ 1e-3, 2e-3\}} & 0.05 & \mr{2}{\{0.0, 0.01, \\0.05, 0.1\}} \\
            &   & AlphaTuning($q{=}2$) & 1e-3 & & 0.0 & \\ \cline{2-7}
            & \mr{5}{GPT\\L} 
                & FT & 1e-4 & \mr{2}{\{1e-4, 2e-4, 5e-4\}} &  0.01 & \mr{2}{\{0.0, 0.01, \\ 0.02\}} \\
            &   & LoRA & 2e-4 & & 0.02 & \\  \cline{3-7}
            &   & AlphaTuning($q{=}3$) & 1e-4 &\mr{3}{\{1e-4, 2e-4, 5e-4,\\ 1e-3, 2e-3\}} & 0.0 & \mr{3}{\{0.0, 0.01, \\0.05, 0.1\}}  \\
            &   & AlphaTuning($q{=}2$) & 1e-4 & & 0.0 & \\
            &   & AlphaTuning($q{=}1$) & 1e-3 & & 0.01 & \\ 
    \Xhline{2\arrayrulewidth}
    \mr{4}{DART \\(Table~\ref{table:DART})}
            & \mr{2}{GPT\\M} 
                & AlphaTuning($q{=}3$) & 1e-3 & \mr{2}{\{1e-4, 2e-4, 5e-4,\\ 1e-3, 2e-3\}} & 0.01 & \mr{2}{\{0.0, 0.01, \\0.05, 0.1\}} \\
            &   & AlphaTuning($q{=}2$) & 1e-3 & & 0.05 & \\ \cline{2-7}
            & \mr{2}{GPT\\L} 
                & AlphaTuning($q{=}3$) & 5e-4 &\mr{2}{\{1e-4, 2e-4, 5e-4,\\ 1e-3, 2e-3\}} & 0.1 & \mr{2}{\{0.0, 0.01, \\0.05, 0.1\}}  \\
            &   & AlphaTuning($q{=}2$) & 1e-3 & & 0.1 & \\
    \Xhline{2\arrayrulewidth}
     \mr{9}{E2E \\(Table~\ref{table:E2E})}
            & \mr{4}{GPT\\M} 
                & FT & 1e-4 & \mr{2}{\{1e-4, 2e-4, 5e-4\}} & 0.02 & \mr{2}{\{0.0, 0.01, \\ 0.02\}} \\
            &   & LoRA & 2e-4 & & 0.01 &\\ \cline{3-7}
            &   & AlphaTuning($q{=}3$) & 2e-3 & \mr{2}{\{1e-4, 2e-4, 5e-4,\\ 1e-3, 2e-3\}} & 0.0 & \mr{2}{\{0.0, 0.01, \\0.05, 0.1\}} \\
            &   & AlphaTuning($q{=}2$) & 5e-3 & & 0.1 & \\ \cline{2-7}
            & \mr{4}{GPT\\L} 
                & FT & 5e-4 & \mr{2}{\{1e-4, 2e-4, 5e-4\}} &  0.01 & \mr{2}{\{0.0, 0.01, \\ 0.02\}} \\
            &   & LoRA & 2e-4 & & 0.0 & \\  \cline{3-7}
            &   & AlphaTuning($q{=}2$) & 1e-3 & \mr{2}{\{1e-4, 2e-4, 5e-4,\\ 1e-3, 2e-3\}} & 0.1 & \mr{2}{\{0.0, 0.01, \\0.05, 0.1\}}  \\
            &   & AlphaTuning($q{=}1$) & 1e-3 & & 0.1 & \\ 
    \Xhline{2\arrayrulewidth}
    \end{tabular}
    \caption{Selected hyper-parameters (learning rates and weight decay) for GPT-M/L results on this paper. We set the hyper-parameter ranges according to the reported parameters in the previous papers \cite{prefixtuning, lora}. For each hyper-parameter selection, the test scores are measured at the last epoch and averaged over 5 trials, which are performed with fixed 5 random seeds.}
    \label{app:table:lrwd}
\end{table*}

\begin{table*}[t]
\small
    \centering
    \begin{tabular}{cccccccc}
    \Xhline{2\arrayrulewidth}
Model & $q$ & Learning rate & Weight decay & Loss & Unseen & Seen & All \\
\Xhline{2\arrayrulewidth}
\mr{8}{GPT-M} 
 & \mr{4}{2}   & 2e-3   & \mr{4}{0.00} & $\boldsymbol{0.84}$ & $36.7_{\pm0.9}$ &  $\boldsymbol{62.7_{\pm0.6}}$ & $51.05_{\pm0.6}$ \\
 &             & 1e-3   & & $0.84$ & $37.3_{\pm0.5}$ & $62.6_{\pm0.5}$ & $\boldsymbol{51.26_{\pm0.5}}$  \\
 &             & 5e-4  & & $0.87$ & $37.8_{\pm0.5}$ & $61.9_{\pm0.5}$ & $51.08_{\pm0.3}$ \\
 &             & 2e-4  & & $0.93$ & $\boldsymbol{38.2_{\pm0.3}}$ & $60.1_{\pm0.2}$ & $50.31_{\pm0.2}$ \\ \cline{2-8}
 & \mr{4}{3}   & 2e-3	&   \mr{4}{0.05}	& $\boldsymbol{0.80}$	& $40.1_{\pm0.6}$ & $\boldsymbol{63.6_{\pm0.2}}$  & $53.1_{\pm0.4}$ \\
 &             & 1e-3	&   	& $0.81$	& $40.9_{\pm0.5}$    & $63.2_{\pm0.5}$  & $\boldsymbol{53.1_{\pm0.4}}$ \\
 &             & 5e-4	&   	& $0.83$	& $\boldsymbol{41.2_{\pm0.4}}$    & $62.7_{\pm0.3}$ & $53.0_{\pm0.3}$ \\ 
 &             & 2e-4	&   	& $0.87$	& $41.0_{\pm0.1}$    & $61.3_{\pm0.2}$  & $52.2_{\pm0.1}$ \\ 
 \Xhline{2\arrayrulewidth}
 \end{tabular}
    \caption{BLEU scores of GPT2-M AlphaTuning on WebNLG dataset when the learning rates vary. Higher learning rates lead to better `Seen' scores, but lead to worse `Unseen' scores (less generative capability). Reversely, lower learning rates lead to better `Unseen' scores.}
    \label{app:table:webnlg-lr}
 \end{table*}

\section{Experimental Details on OPT models}
\label{sec:app:opt_exp}
To study performance on downstream tasks of AlphaTuning using larger PLMs (than GPT-2), we utilize pre-trained OPT models \cite{zhang2022opt} on GLUE-MNLI and SAMSum datasets.
Due to the limitations on resources, our experiments are restrained to 1.3B model\footnote{Available at \url{https://huggingface.co/facebook/opt-1.3b}} with 24 layers ($h$=2048). Fine-tuning and Alphatuning are performed under the conditions that we describe in the following subsections.

\subsection{Dataset}
\textbf{MNLI}\cite{williams-etal-2018-broad}(Multi-Genre Natural Language Inference) evaluates the sentence understanding performance. Given a pair of premise and hypothesis, the main task is to classify the relationship between the two sentences into one of entailment, contradiction, and neutral. A linear classifier head with three output logits is attached on top of the language model and fine-tuned along with the model. The addition of a linear layer slightly increases the overall parameter, unlike other AlphaTuning experiments that only learn $\alpha$.


\textbf{SAMSum}\cite{gliwa2019samsum} is a conversation dataset containing 16k summaries. Given a dialog, the goal is to generate summarizations to evaluate dialog understanding and natural language generation capabilities. For diversity, each conversation style includes informal, semi-formal, and formal types with slang words, emoticons, and typos.

\subsection{Adaptation details}
 Experimental configurations for adaptation are presented in the Table~\ref{app:table:optconfig}. During the training and evaluation of the SAMSum dataset, the beam size is fixed to be 4 and the generation max length is fixed to be 256 (the condition max length is fixed to be 192 and the label max length to be 64). Due to the decoder-only structure of OPT, the token sequence corresponding to the conditions above was put into the input and learned together.
 

\begin{table}[t]
\small
\begin{tabular}{ccccc}
\Xhline{2\arrayrulewidth}
 & \multicolumn{2}{c}{MNLI}                                                                                       & \multicolumn{2}{c}{SAMSum}     \\ \cmidrule(lr){2-3}  \cmidrule(lr){4-5}
Method            & \multicolumn{1}{c}{FT} &  AlphaT                                      & \multicolumn{1}{c}{FT}       & \multicolumn{1}{c}{AlphaT}  \\ 
\Xhline{2\arrayrulewidth}
Learning Rate (LR)     & \multicolumn{1}{c}{5e-6} & \multicolumn{1}{c}{5e-5}  & \multicolumn{1}{c}{6e-6} & \multicolumn{1}{c}{1e-4}   \\ \hline
Weight Decay      & \multicolumn{1}{c}{0.01}     & \multicolumn{1}{c}{0.05}     & \multicolumn{1}{c}{0.01}     & \multicolumn{1}{c}{0.05}      \\ \hline
Optimizer         & \multicolumn{2}{c}{AdamW}                                                                                                       & \multicolumn{2}{c}{Adafactor}                                                                                \\ \hline
Epoch             & \multicolumn{2}{c}{3}                                                                                                           & \multicolumn{2}{c}{5}            \\ \hline                        
LR Scheduler         & \multicolumn{4}{c}{Linear decay}                               \\ \hline
Batch             & \multicolumn{4}{c}{32}                                                                                        \\ \hline
\Xhline{2\arrayrulewidth}
\end{tabular}
\caption{Hyper-parameter selection for the experiments using OPT models}
\label{app:table:optconfig}
\end{table}

\section{Details on BCQ Format}
This section introduces two popular methods to produce binary codes and scaling factors from full-precision DNN weights. 
The common objective is to minimize mean square error (MSE) between original data and quantized data in heuristic manner.
As introduced in \cite{rastegariECCV16} (\ie $q$=1), a weight vector $\vw$ is approximated to $\alpha\vb$ where $\alpha$ is a full-precision scaling factor and $\vb$ is a binary vector ($\vb \in \{-1, +1\}^n$).
For one-bit quantization, there is an analytic solution to minimize $\Arrowvert \vw-\alpha\vb\Arrowvert^2$ as following:

\begin{equation} \label{eq:ap:1}
\vb^* = {\rm sign}(\vw),\; \alpha^* = \frac{\vw^\top \vb^*}{n}.
\end{equation}

However, if we extend this equation to multi-bit ($q>1$) quantization, there is no analytic solution.

\textbf{Greedy Approximation}
first produces $\alpha_1$ and $\vb_1$ as in Eq.~\ref{eq:ap:1}, and then calculates $\alpha_i$ and $\vb_i$ iteratively by minimizing the residual errors ($\Arrowvert \vw-\sum_{j=1}^{i-1} \alpha_j\vb_j\Arrowvert^2$).
Then, $\alpha_i$ and $\vb_i$ are calculated as follows:

\begin{equation} \label{eq:ap:2}
\vb_i^* = {\rm sign}(\vw-\sum_{j=1}^{i-1} \alpha_j\vb_j).
\end{equation}
\begin{equation} \label{eq:ap:3}
\alpha_i^* = \frac{\vw-\sum_{j=1}^{i-1} \alpha_j\vb_j^\top \vb_i^*}{n}.
\end{equation}

Although this method is computationally simple, it leads to higher MSE by quantization.
In spite of higher quantization error, AlphaTuning utilizes this Greedy method only for the initial PTQ process. 
We observe the adapted LMs with AlphaTuning (along with Greedy approximation) can reach the comparable scores of full fine-tuning or LoRA while there is no noticeable improvement by using the Alternating method, an advanced method that we discuss next.

\textbf{Alternating Method} \cite{xu2018alternating} adjusts scaling factors and binary values iteratively after producing the initial $\alpha_{1..q}$ and $\vb_{1..q}$ obtained by Greedy approximation method.
From the initial $\vb_{1..q}$, $\alpha_{1..q}$ can be refined as
\begin{equation} \label{eq:ap:4}
\left[ \alpha_1, ..., \alpha_q \right] = \left( \left( \mB_q^\top \mB_q \right)^{-1} \mB_q^\top \vw \right)^\top,
\end{equation}
where $\mB_q = \left[ \vb_1, ... ,\vb_q \right] \in \{-1,+1\}^{n \times q}$.
From the refined $\alpha_{1..q}$, the elements in $\vb_{1..q}$ can be further refined using a binary search algorithm.
As we iterate the process of refining the scaling factor and the binary vector repeatedly, the errors due to quantization get reduced.
When the amount of error is reduced to become close enough to zero, we can stop such iterative processes.
It has been known that an appropriate number of iterations is approximately between 3 and 15 \cite{xu2018alternating, quant_lee}, and this paper set the number of iterations as 15 when the Alternating method is selected for PTQ process.

In practice, higher scores right after PTQ are attainable by the Alternating quantization method rather than the Greedy method.
Thus, we try previous experiments using Alternating method as shown in Table~\ref{app:table:webnlgptq} and \ref{app:table:e2eptq} on WebNLG and E2E dataset.
From those two tables, it should be noted that even when the Alternating algorithm is chosen, we can observe that post-training quantization still leads to considerable performance degradation.

\textbf{group-wise quantization}
While Eq.~\ref{eq:ap:1}-\ref{eq:ap:3} are described for a weight `vector' ($\vw$) (for simplicity),
the target parameters to be quantized in this paper are in the form of weight `matrices' of LMs.
We extended the principles of weight vector quantization to a row-wise quantization scheme of a weight matrix (\ie for each row, $q$ of scaling factors are produced) in Eq.~\ref{eq:1}.
In this case, $g$ should be set to $h_{in}$ as described in Table~\ref{table:quant_example}.
If we assign the $g$ to be $h$ (\ie, a hidden size of a model), each row will be divided into $h_{in}/h$ of vectors, and each divided vector will produce its own scaling factors.
Although we did not explain implementation issues of such a group-wise quantization, it has been also shown that group-wise quantization has minimal impact on inference latency \cite{sc22}.

\begin{table*}[ht]
\small
    \centering
    \begin{tabular}{clccccc}
        \Xhline{2\arrayrulewidth}
        Model & Method & $q$ & Loss & Unseen & Seen & All \\
        \Xhline{2\arrayrulewidth}
        \mr{7}{GPT\\M} 
                  & \mr{3}{\PTQA}  &2 &5.57 & $4.9_{\pm.8}$ & $13.9_{\pm2.1}$ & $10.1_{\pm1.7}$\\
                & & 3 & 2.03 & $25.0_{\pm2.5}$ & $58.7_{\pm1.0}$ & $43.2_{\pm3.3}$\\
                & & 4 & 1.70 & $30.6_{\pm.9}$ & $63.3_{\pm.3}$ & $47.8_{\pm.7}$\\ \cline{2-7}
                & \mr{2}{\PTQB} & 3 & 2.98 & $15.8_{\pm3.0}$ & $15.8_{\pm3.4}$ & $15.8_{\pm3.2}$ \\
                & & 4 & 2.72 & $19.8_{\pm2.4}$ & $23.9_{\pm3.7}$ & $22.2_{\pm3.1}$ \\ \cline{2-7}
                & \mr{2}{\PTQC } & 3 & 3.36 & $12.6_{\pm4.1}$ & $16.6_{\pm6.7}$ & $13.6_{\pm7.5}$\\
                & & 4 & 3.10 & $13.4_{\pm5.1}$ & $15.9_{\pm6.9}$ & $14.4_{\pm6.0}$\\
        \Xhline{2\arrayrulewidth}
        \mr{7}{GPT\\L} 
                  & \mr{3}{\PTQA} & 2 & 2.11 & $19.5_{\pm1.9}$ & $62.2_{\pm0.2}$ & $41.2_{\pm2.0}$\\
                & & 3 & 1.91 & $23.2_{\pm.8}$ & $62.7_{\pm.2}$ & $43.7_{\pm.7}$\\
                & & 4 & 1.87 & $23.7_{\pm.3}$ & $61.8_{\pm.4}$ & $43.6_{\pm.3}$\\ \cline{2-7}
                & \mr{2}{\PTQB} & 3 & 1.97 & $20.1_{\pm5.2}$ & $27.8_{\pm4.1}$ & $24.1_{\pm4.5}$ \\
                & & 4 & 1.67 & $33.9_{\pm4.4}$ & $45.9_{\pm3.7}$ & $41.7_{\pm2.1}$ \\ \cline{2-7}
                &\mr{2}{\PTQC} & 3 & 1.97 & $14.0_{\pm7.2}$ & $26.6_{\pm11.5}$ & $25.8_{\pm13.0}$\\
                & & 4 & 1.67 & $28.7_{\pm7.1}$ & $34.3_{\pm13.1}$ & $29.9_{\pm12.1}$\\
        \Xhline{2\arrayrulewidth}
    \end{tabular}
    \caption{BLEU scores on WebNLG dataset with post-training quantization. The fine-tuned models (with W\textsubscript{FT}) and LoRA-tuned models (with frozen W and W\textsubscript{LoRA}) are quantized by Alternating method \cite{xu2018alternating} without gradient updates (post-training quantization).}
    \label{app:table:webnlgptq}
\end{table*}

\begin{table*}[ht]
\small
    \centering
    \begin{tabular}{clccccccc}
        \Xhline{2\arrayrulewidth}
        Model & Method & $q$ & loss & BLEU & NIST & METEOR & ROUGE\_L & CIDEr \\
        \Xhline{2\arrayrulewidth}
        \mr{7}{GPT\\M} 
                  & \mr{3}{\PTQA} & 2 & 2.687 & $50.2_{\pm2.2}$ & $5.78_{\pm1.2}$ & $35.8_{\pm1.5}$ & $60.1_{\pm.9}$ &$1.42_{\pm.11}$\\
                & & 3 & 1.192 & $67.5_{\pm.5}$ & $8.58_{\pm.04}$ & $46.3_{\pm.5}$ & $70.3_{\pm.1}$ & $2.39_{\pm.01}$\\
                & & 4 & 0.992	& $67.2_{\pm.6}$ & $8.52_{\pm.07}$ & $46.4_{\pm.3}$ & $70.3_{\pm.4}$	 & $2.38_{\pm.03}$\\ \cline{2-9}
                & \mr{2}{\PTQB} &3 & 4.095 & $11.1_{\pm5.3}$ & $2.35_{\pm1.04}$ & $12.4_{\pm4.2}$ & $29.9_{\pm7.8}$ & $0.35_{\pm.1}$\\
                & & 4 & 1.916 & $48.3_{\pm3.8}$ & $3.65_{\pm1.16}$ & $32.3_{\pm.97}$ & $59.8_{\pm1.8}$ & $1.22_{\pm.14}$ \\ \cline{2-9}
                & \mr{2}{\PTQC } & 3 & 4.377 & $7.5_{\pm3.2}$ & $1.31_{\pm.21}$ & $11.4_{\pm.8}$ & $29.8_{\pm3.0}$ & $0.29_{\pm.04}$\\
                & & 4 & 3.561 & $14.6_{\pm4.8}$ & $1.41_{\pm1.1}$ & $15.7_{\pm1.2}$ & $38.7_{\pm3.2}$ & $0.43_{\pm.03}$\\
        \Xhline{2\arrayrulewidth}
        \mr{7}{GPT\\L} 
                  & \mr{3}{\PTQA} & 2 & 0.998 & $66.1 _{\pm1.0}$ & $8.40_{\pm.11}$ & $45.5_{\pm.4}$ & $70.0_{\pm.4}$ & $2.33_{\pm.03}$\\
                & & 3 & 0.979 & $66.5_{\pm1.0}$ & $8.45_{\pm.10}$ & $45.7_{\pm.3}$ & $70.3_{\pm.5}$ & $2.37_{\pm.03}$\\
                & & 4 & 0.976 & $66.6_{\pm1.0}$ & $8.47_{\pm.09}$ & $45.8_{\pm.2}$ & $70.3_{\pm.4}$ & $2.37_{\pm.01}$\\ \cline{2-9}
                &\mr{2}{\PTQB} &3 &1.868 & $50.9_{\pm3.4}$ & $6.63_{\pm.32}$ &$38.7_{\pm1.9}$ & $60.6_{\pm1.9}$ & $1.30_{\pm.19}$\\
                & & 4 & 1.398 & $65.4_{\pm2.0}$ & $8.46_{\pm0.19}$ & $44.9_{\pm.6}$ & $65.3_{\pm2.2}$ & $2.15_{\pm.11}$\\\cline{2-9}
                & \mr{2}{\PTQC } & 3 & 4.377 & $7.5_{\pm3.2}$ & $1.31_{\pm.21}$ & $11.4_{\pm.8}$ & $29.8_{\pm3.0}$ & $0.29_{\pm.04}$\\
                & & 4 & 3.561 & $14.6_{\pm4.8}$ & $1.41_{\pm1.1}$ & $15.7_{\pm1.2}$ & $38.7_{\pm3.2}$ & $0.43_{\pm.03}$\\
        \Xhline{2\arrayrulewidth}
    \end{tabular}
    \caption{Test scores on E2E dataset after post-training quantization ($q=3$) performed by Alternating method.}
    \label{app:table:e2eptq}
\end{table*}

\begin{table*}[t]
\small
\centering
\begin{threeparttable}
\begin{tabular}{ccccccccccccc}
\Xhline{2\arrayrulewidth}
\mr{2}{Model}       & \mr{2}{Method} & \mr{2}{q}  & \mr{2}{Trainable \\ Params}      & \multicolumn{9}{c}{GLUE}        \\ \cline{5-13} 
                    &  &  & &  CoLA  & SST-2 & MRPC  & STS-B & QQP   & MNLI \tnote{1} & MNLI\textsubscript{mm} \tnote{2}& QNLI  &  RTE  \\ 
\Xhline{2\arrayrulewidth}
\mr{2}{Base}  & Fine-Tuning  & - & 108.3M & 52.1  & 93.5  & 88.9  & 85.8 & 71.2  & 84.6 & 83.4    & 90.5 &  66.4 \\ \cline{2-13} 
                    & AlphaTuning  & 3 & 0.1M & 51.0  & 91.4  & 91.4  & 87.4 & 84.2  & 80.8 & 81.1    & 89.4 &  69.3 \\ \hline
\mr{3}{Large}& Fine-Tuning  & - & 333.6M & 60.5  & 94.9  & 89.3  & 86.5 & 72.1  & 86.7 & 85.9    & 92.7  &  70.1  \\ \cline{2-13} 
                    & AlphaTuning  & 2 & 0.3M & 49.1  & 90.9  & 88.8  & 87.0   & 84.9  & 82.7 & 83.5    & 89.9  &  66.8  \\ \cline{2-13} 
                    & AlphaTuning  & 3 & 0.3M & 55.7  & 92.3  & 88.9  & 87.9 & 85.6  & 83.8 & 84.7    & 91.4  &  63.2  \\ 
\Xhline{2\arrayrulewidth}
\end{tabular}
    \begin{tablenotes}
        \item[1] MNLI-matched
        \item[2] MNLI-mismatched
  \end{tablenotes}
   \end{threeparttable}
\caption{BERT-base-cased and BERT-large-cased with full fine-tuning and AlphaTuning. Experiments were evaluated on the GLUE benchmark \cite{wang2018glue}. The fine-tuning results used for comparison refer to \cite{BERT}. AlphaTuning follows the accuracy of full fine tuning. The results show that AlphaTuning can be applied to the BERT architecture. Experimental details are in Table~\ref{app:table:bertconfig}.}
\end{table*}

\begin{table*}[t]
\small
\centering
\begin{threeparttable}
\begin{tabular}{ccccccccccc}
\Xhline{2\arrayrulewidth}
\mr{2}{Model}       & \mr{2}{Configuration}   & \multicolumn{9}{c}{GLUE}        \\ \cline{3-11} 
                    & &  CoLA  & SST-2 & MRPC  & STS-B & QQP   & MNLI \tnote{1} & MNLI\textsubscript{mm} \tnote{2}& QNLI  &  RTE  \\ 
\Xhline{2\arrayrulewidth}
\mr{2}{Base}  & Batch size & 16  & 32 & 32  & 32 &32  & 16 & 16    & 16 &  16 \\ \cline{2-11} 
                    & Learning rate  & 1e-4  & 1e-4  & 1e-4  & 2e-4& 1e-4  & 5e-5 & 5e-5    & 5e-5 & 1e-4 \\ \hline
\mr{3}{Large}& Batch size   & 32 & 16  & 16 & 16 & 32 &16 & 16   & 16 &  16  \\ \cline{2-11} 
                    & Learning rate  & 1e-4  & 1e-4  & 1e-4  & 1e-4   & 5e-5  & 5e-5 & 5e-5    & 5e-5  &  1e-4  \\ 
\Xhline{2\arrayrulewidth}
\end{tabular}
    \begin{tablenotes}
        \item[1] MNLI-matched
        \item[2] MNLI-mismatched
  \end{tablenotes}
   \end{threeparttable}
\caption{Hyper-parameter selection for the experiments using BERT-base and BERT-large on GLUE benchmark. For each experiment, the optimizer is selected to be AdamW with a linear decaying learning rate scheduler. The number of epochs is fixed to be 3, and the weight decay is set to 0.01. The metrics used in the evaluation are Matthew’s correlation for CoLA, Pearson correlation for STS-B, F1 score for QQP and MRPC and accuracy for the other tasks.}
\label{app:table:bertconfig}
\end{table*}

\end{document}